\documentclass[10pt,twocolumn,letterpaper]{article}

\usepackage{wacv}
\usepackage{times}
\usepackage{epsfig}
\usepackage{graphicx}
\usepackage{amsmath}
\usepackage{amssymb}
\usepackage{bm}
\usepackage{booktabs} 
\usepackage{multirow}
\usepackage[table,xcdrsaw]{xcolor}
\graphicspath{{figures/}}
\usepackage{subcaption}
\usepackage{url}
\usepackage[title]{appendix}
\usepackage[font=small,skip=0pt]{caption}  
\usepackage{stfloats}
\usepackage{xcolor}  
\usepackage{soul}  
\setstcolor{red}

\wacvfinalcopy 

\def\wacvPaperID{740} 

\ifwacvfinal\fi
\begin{document}

\title{Measuring the Utilization of Public Open Spaces by Deep Learning: a Benchmark Study at the Detroit Riverfront}

\author{Peng Sun \hspace{2cm} Rui Hou \hspace{2cm} Jerome P. Lynch\\
University of Michigan\\
{\tt\small $\{$patcivil, rayhou, jerlynch$\}$ @umich.edu}
}

\maketitle
\ifwacvfinal\thispagestyle{empty}\fi

\begin{abstract}
Physical activities and social interactions are essential activities that ensure a healthy lifestyle. Public open spaces (POS), such as parks, plazas and greenways, are key environments that encourage those activities. To evaluate a POS, there is a need to study how humans use the facilities within it. However, traditional approaches to studying use of POS are manual and therefore time and labor intensive. They also may only provide qualitative insights. It is appealing to make use of surveillance cameras and to extract user-related information through computer vision. This paper proposes a proof-of-concept deep learning computer vision framework for measuring human activities quantitatively in POS and demonstrates a case study of the proposed framework using the Detroit Riverfront Conservancy (DRFC) surveillance camera network. A custom image dataset is presented to train the framework; the dataset includes 7826 fully annotated images collected from 18 cameras across the DRFC park space under various illumination conditions. Dataset analysis is also provided as well as a baseline model for one-step user localization and activity recognition. The mAP results are 77.5\% for {\it pedestrian} detection and 81.6\% for {\it cyclist} detection. Behavioral maps are autonomously generated by the framework to locate different POS users and the average error for behavioral localization is within 10 cm.   
\end{abstract}

\section{Introduction}
\label{section:introduction}
POS are a vital part of healty cities offering public spaces for social interactions, exercise and to enjoy with nature.  The global trend of shrinking households drives a need for social contact outside the home \cite{gehl2013cities}. Healthy living \cite{arena2017HealthPark} encourages people to stay physically active within a pleasant environment (e.g. green parks). Studies \cite{sallis2012role,francis2012quality,giles2015translating} also show that physical activities which can be promoted in POS are beneficial to mental health and can substantially reduce the risk of chronic disease (e.g. cardiovascular disease, pulmonary disease, metabolic syndromes).  People-centered urban design of POS is drawing increasing attention with designed POS to promote physical activities \cite{mccormack2010characteristics} with post occupancy evaluation (POE) \cite{preiser2015poe} later performed to verify design assumptions. Methods of measuring the usage of POS (e.g. counting, mapping, tracing, and test walks) are often executed manually, which are time and labor intensive \cite{gehl2013study}. Hence, an automatic sensing method is needed to investigate patrons within POS. 

    \begin{figure*} [tp]
        \vspace{-1.8em}   
        \centering
        \begin{subfigure}[b]{0.51\textwidth}
            \centering
            \includegraphics[width=\textwidth]{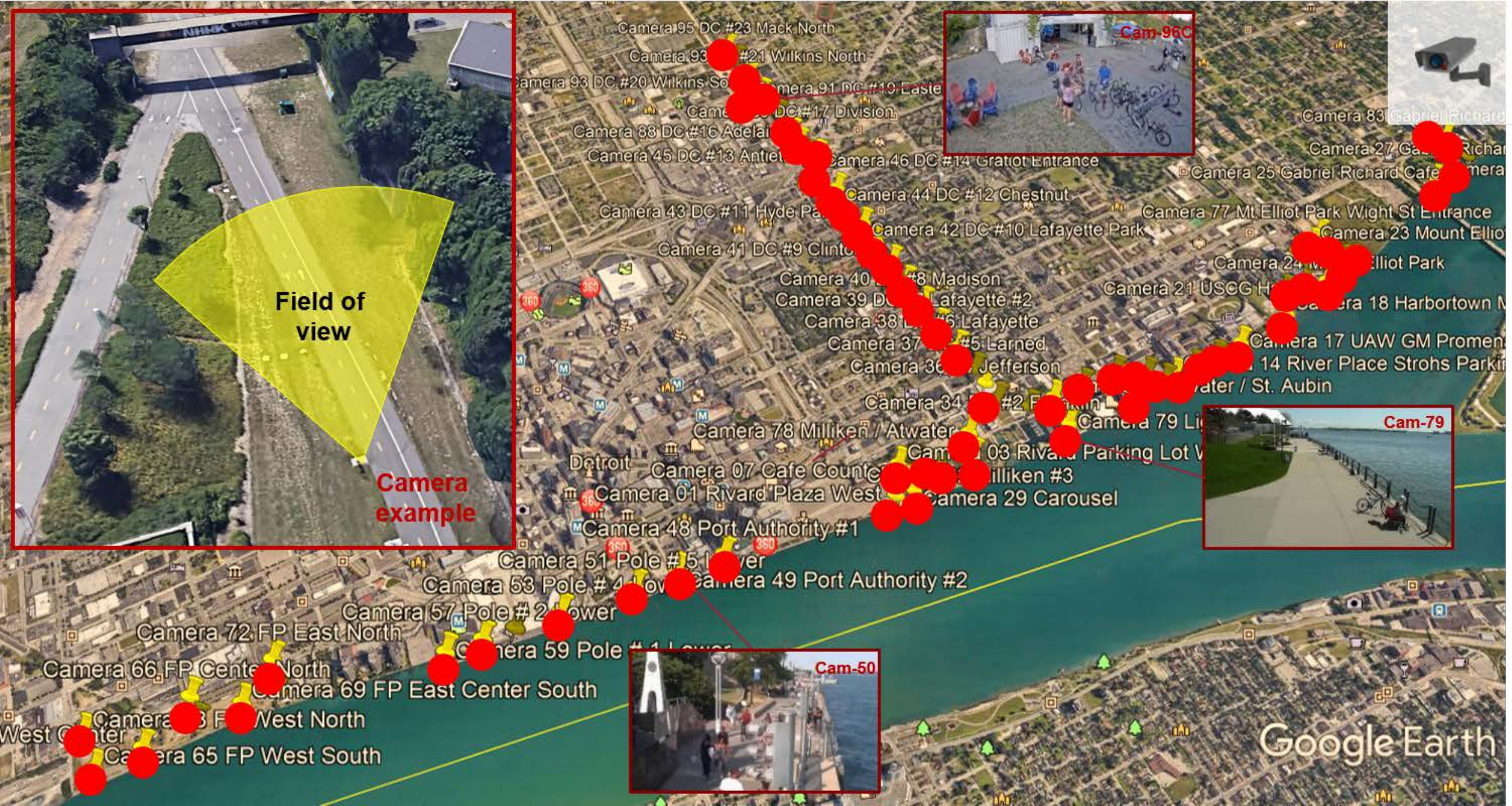}
            \caption[]%
            {{\small }}    
            \label{fig:DRC_map}
        \end{subfigure}
        \quad
        \begin{subfigure}[b]{0.42\textwidth}  
            \centering 
            \includegraphics[width=\textwidth]{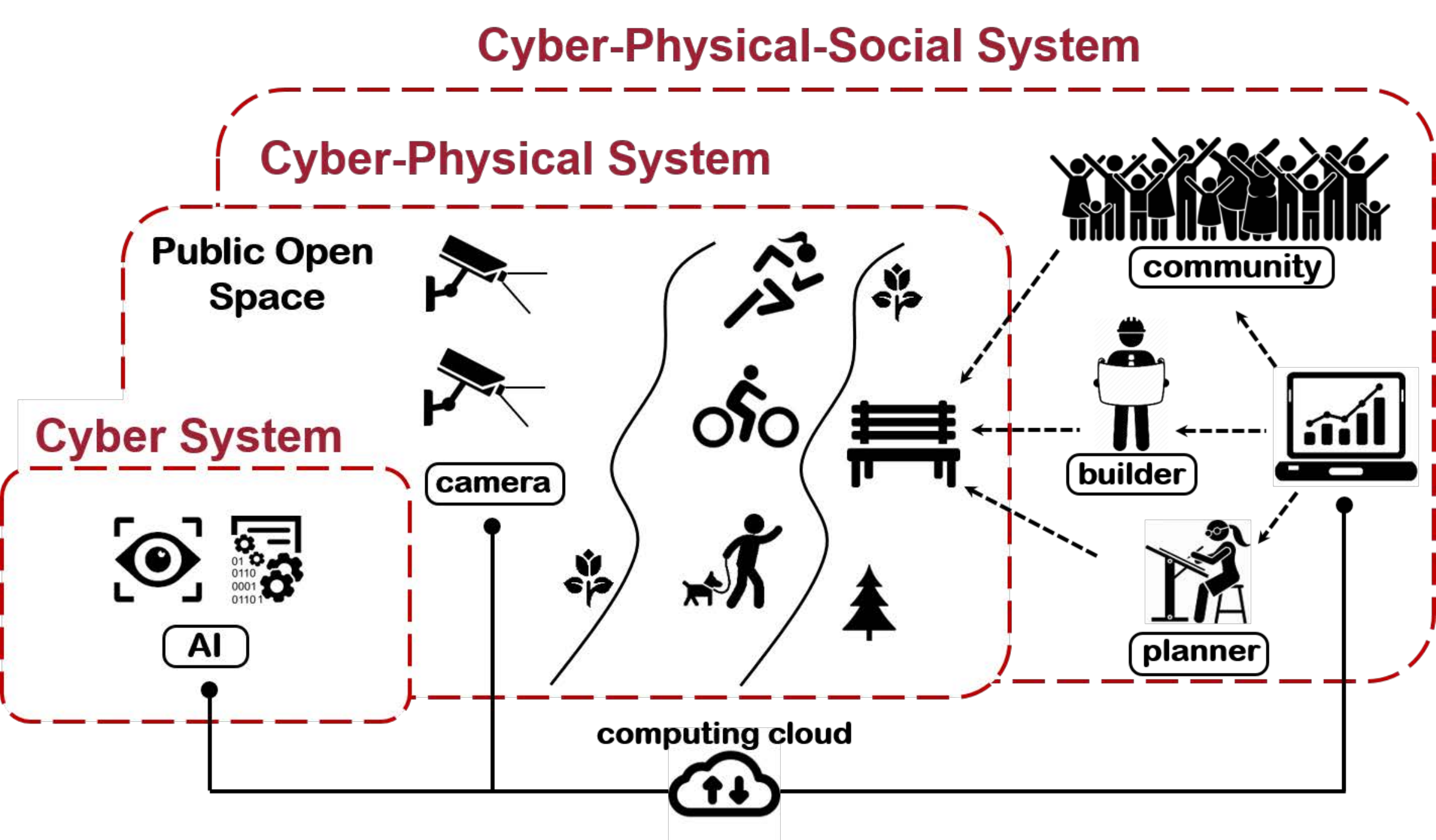}
            \caption[]%
            {{\small }}    
            \label{fig:cv_cps}
        \end{subfigure} \\
        \caption{(a) A map of the Detroit riverfront with surveillance cameras. (b) CV-based cyber-phsyical-social system (CPSS) for urban planning and design.} 
        \vspace{-1.8em}   
    \end{figure*}
    
POS (e.g. parks and greenways) can serve as anchoring points driving the transformation of urban spaces in populated cities into more lively environments. In 2003, the Detroit Riverfront Conservancy (DRFC) was incorporated to restore the international riverfront area of Detroit (Fig. \ref{fig:DRC_map}). A network of 100 surveillance cameras (Fig. \ref{fig:DRC_map}) has been installed at the riverfront to ensure the safety of the area.  These surveillance cameras provide a means of automating the assessment of patrons within POS using computer vision (CV).  Recently, deep learning based-CV techniques have benefited many domains including video surveillance, assisted living, human-robot interaction, and health care. This paper presents a CV-based approach to automatically localize users and recognize their activities for measuring the usage of POS using surveillance cameras. The proposed cyber-physical-social system (Fig. \ref{fig:cv_cps}) can provide informative guidelines for improving urban spaces during their design. Users' activity information (e.g. usage patterns and occupancy frequencies) is extracted by CV algorithms and shared with urban designers and park managers for decision making.  

The paper is organized as follows:
first, a review of the existing related works on vision-based urban studies is presented; second, the detection model, behavioral mapping, and the evaluation metrics are presented; third, the collection and annotation process of a custom dataset is described; fourth, a case study of using the proposed framework to study the usage of a POS at the Detroit Riverfront is presented.  The key contributions of this study are in three-fold: (1) a custom image dataset is established for user detection in POS as well as evaluation metrics, (2) a baseline model for POS sensing is trained and evaluated for user localization and activity recognition using monocular surveillance cameras, and (3) a benchmark study is demonstrated and heat-maps of user behaviors are depicted for key park locations.

\section{Related Work}
\label{section:relatedwork}
{\bf Vision-based sensing in urban studies:} Architects started to adopt video recordings to study people's social behavior and interactions in public spaces since the late 1960's \cite{whyte1980social}. Manual observation is the primary way for such a purpose. With the development of CV-based sensing technology, coupled with cheap computing power, there are emerging studies using CV to measure physical urban changes \cite{naik2017urbanchange}, to analyze pedestrian gait \cite{hediyeh2014pedestrian}, and to study autonomous driving \cite{ barnes2017urbanauto, hoermann2018dynamic} in urban streets. However, few studies have been focused on measuring the usage of POS for improving urban planning. \cite{yan2005public} is one of the few studies that employed a computer vision-based method to measure human activity (e.g. sitting or walking) in a POS. The people detection method was based on background subtraction and blob detection.  Although it introduced video sensing into urban space studies, the sensing system was unable to perform real-time detection; the detection robustness and accuracy is also presumed to suffer from adopting low-level features of images. An advanced detection method should be investigated to perform real-time, robust user sensing in POS by using deep features of images.

{\bf CV-based object detection:} Object detection models are utilized to identify and locate objects in images. Region-based detection models (e.g. Fast R-CNN \cite{girshick2015fast} and Faster R-CNN \cite{ren2015faster}) rely on region proposal networks (RPN) \cite{ren2015faster} and convolutional neural networks (CNN) to estimate bounding boxes (bbox) of objects. In contrast, single-stage models (e.g. YOLO \cite{redmon2016yolo} and SSD \cite{liu2016ssd}) perform object detection without a separate region proposal step.  Although the former methods suffer from comparatively slow detection speed, they outperform the latter in detection accuracy\cite{zhao2019ObjDetReview}.  Mask R-CNN \cite{he2017mask} is a region-based detection method that provides richer information of a detected object with an additional instance segmentation besides a bbox. Furthermore, detected contours can provide location information of specific body parts \cite{li2017humanParsing}. Recently, new anchor-free detectors (e.g. FCOS \cite{tian2019fcos}, FoveaBox \cite{kong2019foveabox}) have been developed to achieve higher performance in detecting accurate bounding boxes without using anchor references. 

\begin{table*}[tp]
\footnotesize
\centering
\vspace{-1.5em}   
\begin{tabular}{@{}p{1.5cm}p{1.6cm}p{0.8cm}p{1.8cm}p{1.1cm}p{.7cm}p{2.4cm}p{0.6cm}p{0.6cm}p{1.7cm}@{}}
\toprule
Dataset                                               & Main purpose   & \# Cls & \# Ppl. cls & \# Annt. imgs & Inst. segm. & Cam/sensor type                                                   & Night & Rain & Locations \\
\midrule
ImageNet-1k                                                    & object det.             & 1000            & 1 (pers.)            & 1.3M                 & No                   & -                                                                          & No             & No            & -                  \\
VOC-2007                                                       & object det.             & 20              & 1 (pers.)            & 10K                   & No                   & -                                                                          & No             & No            & -                  \\
COCO-2014                                                      & object det.             & 80              & 1 (pers.)            & 160K                  & Yes                  & -                                                                          & No             & No            & -                  \\
CityScapes                                                     & auton. driving          & 19              & 2 (pers.+rider)     & 25K                   & Yes                  & \begin{tabular}[t]{@{}l@{}}veh.-mount. \\     dash cam.\end{tabular}       & No             & No            & 50 cities in Europe  \\
\begin{tabular}[c]{@{}l@{}}Mapillary\\   Vistas\end{tabular}   & auton. driving          & 66              & 5 (pers.+rider)      & 25K                   & Yes                  & phone, tablet, action cams, etc.                                           & No             & No            & around world       \\
nuScenes                                                       & auton. driving          & 23              & 4 (ped. etc.)          & 1.4M                   & Yes                  & \begin{tabular}[t]{@{}l@{}}veh.-mount.cam.\\     +lidar+ladar\end{tabular} & Yes            & Yes           & Boston, SG         \\
\begin{tabular}[c]{@{}l@{}}Caltech\\   Pedestrian\end{tabular} & ped. det.               & 1               & 1 (ped.)             & 250k                  & No                   & \begin{tabular}[t]{@{}l@{}}veh.-mount. \\     dash cam.\end{tabular}       & No             & No            & LA       \\
\begin{tabular}[c]{@{}l@{}}Cyclist\\   Benchmark\end{tabular}  & cyclist det.            & 1               & 1 (cycl.)            & 14.6K                 & No                   & \begin{tabular}[t]{@{}l@{}}veh.-mount.\\     stereo cam.\end{tabular}      & No             & No            & Beijing            \\
OPOS                                                     & user sense  & 15              & 10                   & 7.8K                  & Yes                  & PTZ surv. cam.                                                             & Yes            & Yes           & Detroit    \\
\bottomrule    
\end{tabular}
\caption{Comparison of image datasets for object detection.}
\label{tab:datasets}
\vspace{-2 em}   
\end{table*}
\normalsize

{\bf Human activity recognition:} For many years, researchers have been working on human activity recognition (HAR) using different kinds of sensors. Non-vision based (e.g. wearable) and vision-based sensors are two categories of sensors used. In \cite{lara2012wearable, yang2008distributed}, multiple wearable sensors (e.g. accelerometers, gyroscopes, and magnetometers) are attached to the body of a subject to measure motion attributes in order to recognize different activities. However, wearable sensors are intrusive to users \cite{jalal2017depthvideo} and can only cover a very limited number of users in a POS. Traditional CV methods using vision-based sensors usually rely on a few visual features (e.g. HoG, local binary pattern, or RGB), which makes it difficult to achieve robust people detection (especially under extreme illumination conditions). In the past few years, deep learning \cite{lecun2015deep, goodfellow2016deep} methods using deep neural networks have grown rapidly and are drawing much attention as the result of their supreme performance in different applications.  Others have used deep features of images to extract high-level representation for activity recognition tasks \cite{caba2015activitynet,liu2018multi,gkioxari2018interaction}. A complete HAR study \cite{vrigkas2015HARreview} usually consists of at least two steps: 1) person detection, and 2) activity/body shape recognition based on feature representations (e.g. silhouette representation). Unlike past HAR studies, the objective of this study is to achieve person detection and activity recognition in a single step. To do so, the recognition of different activities will be embedded within the classification task of the instance segmentation.

{\bf Image datasets related to people detection:} Image datasets have promoted research in many topics in computer vision, such as object detection, pedestrian detection, and autonomous driving.  Among the datasets available for generic object detection (Table \ref{tab:datasets}), ImageNet \cite{deng2009imagenet}, COCO \cite{lin2014microsoftcoco}, and PASCAL VOC \cite{everingham2010pascal} are very popular choices with many object classes associated with each of them.  However, there are only a few ``person'' classes in these datasets which limit human activity classification.  The earlier datasets  which were designed solely for pedestrian detection (e.g. Caltech Pedestrian \cite{dollar2009pedestrian} and INRIA Person \cite{dalal2005inria}) only include pedestrians.  The human pose in these datasets is either standing straight or walking, while cyclists are excluded or ignored on the annotated images to avoid dealing with the similar appearance between pedestrians and cyclists.  In 2016, a cyclist benchmark dataset \cite{li2016cyclist} was designed for cyclist detection. However, there is only one object class (cyclist) annotated in bbox without any instance segmentations.  Today, because of the emerging technologies of autonomous driving, many datasets are built from data collected from diverse set of sensors mounted on vehicles (e.g. cameras, radar, GPS, IMU, and LIDAR sensors).  Because the main focus for autonomous driving is on the quick detection of pedestrians and other vehicles on the road during driving, the related datasets include only a few classes of people on the road.  For example, CityScapes \cite{cordts2016cityscapes} includes pedestrian and rider, Mapillary Vistas \cite{neuhold2017mapillary} includes persons (individual and groups) and riders (motorcyclist, bicyclist, other), and nuScenes \cite{caesar2019nuscenes} includes different pedestrians (adult, child, construction worker, and police officer).  Hence, in order to detect user activities that are usually observed in a POS (e.g. walking, riding a bike, riding a scooter, sitting), the aforementioned datasets would not achieve the joint task of user detection and activity recognition. A new dataset, including people with various physical activities (that often occur in POS), is needed to train a detection model for POS.

\section{Methodology}
\label{section:method}
\subsection{Detection Model}
\label{section:method_model}

A detection model serves as the primary building block of the sensing framework proposed herein for modeling public use of city park spaces. A user mapping algorithm (mapping from the 2D pixel coordinate system  of the camera to the 3D coordinate system of the real physical space) is built on top of the detection model to assess POS utilization.  In this study, the Mask R-CNN \cite{he2017mask} model is utilized for user detection and activity recognition.  In Mask R-CNN, different CNN architectures can be adopted as the computational core to extract features from an input image. ResNet \cite{he2016deep} is a well-known convolutional backbone to achieve object detection and segmentation. ResNet extracts multiple feature maps from an image while RPN generates regions within which objects potentially lie. The RPN slides over either one of the last convolutional feature maps or over feature maps combined during the use of a feature pyramid network (FPN) \cite{lin2017feature}. FPN serves as a feature detector on top of the feature maps generated by ResNet to gather rich information (i.e. multi-scale features) in both semantics and location. RPN will be trained together with the ResNet backbone in an end-to-end fashion.

\subsection{Detection Evaluation Metrics}
\label{section:method_metrics}
The accuracy of a detector can be evaluated using average precision (AP) and average recall (AR) under a certain threshold of intersection over union (IoU). IoU measures the percentage of overlap between the ground truth and the detected bounding box. In this study, AP and AR are defined following the COCO metrics \cite{lin2014microsoftcoco} which are averaged over multiple IoU values from 0.5 to 0.95 with a step size of 0.05. The mean average precision (mAP) is defined as the mean value of the AP$_c$ across different object classes, $c$.  For a specific class, $c$, AP is also divided into AP$_c^{sm}$ (small), AP$_c^{med}$ (medium), and AP$_c^{lg}$ (large), depending on sizes of objects within images. In order to analyze detection performance and detection error, a single AP cannot provide a thorough diagnosis for a trained detector.  For instance, the effects of different types of false positives \cite{hoiem2012diagnosing} may include the localization error and confusion with semantically similar objects. Hence, the popular precision-recall (PR) curve methods \cite{buckland1994PR, davis2006PRvsROC} will also be utilized in this study to diagnose possible detection errors.

\subsection{Camera Calibration}
\label{section:method_mapping}
The pinhole camera model \cite{ma2012imagemodel} is used to perform camera calibration and 3D bbox estimation \cite{mousavian2017bbox} using geometry. In this work, a scene view is formed by projecting points of a 3D location defined in world coordinates $\{X, Y, Z\}$ onto the image plane defined by the pixel coordinate system $\{u, v\}$ using a perspective transformation:

\begin{equation}
	s\bm{m'}=\bm{A \left[ R|t \right] M} 
	\label{eq:cam_matrix} 
\end{equation}

\noindent where $\bm{A}$ is a matrix of intrinsic parameters of the camera, $[\bm{R|t}]$ is the joint rotation-translation matrix (including extrinsic parameters), and $\bm{m'}$ and $\bm{M}$ are the locations in the pixel coordinate system and world coordinate system, respectively.  The intrinsic matrix $\bm{A}$ and distortion coefficients (i.e. radial and tangential distortion) can be computed by using a chessboard at various positions.  The extrinsic parameters of a camera can be computed by the Levenberg--Marquardt algorithm \cite{levenberg1944method} relating the locations of objects in both world and pixel coordinates (e.g. the pattern of the broken white lines on the pedestrian path).

\subsection{Behavioral Mapping and Evaluation}
\label{section:method_behavior}
Behavioral mapping \cite{sanoff1971behavioral} is a standard method in urban design used to evaluate how participants utilize a built environment by recording the behaviours of participants.  It can provide a useful record of where people are, what they do, and how their behaviors are distributed in space. Behavioral maps consist of two forms: place-centered and individual-centered.\cite{sommer1997practical}  In this paper, the former type of behavioral mapping is studied and monocular surveillance cameras are used to estimate the 3D bbox and the location of users in a POS. The ground is presumed to be flat and the origin, $O$, is fixed to the ground plane ($Z=0$).  This restraint changes the once ill-posed problem into an optimization problem. The mapping algorithm is implemented on top of the Mask R-CNN detector allowing detection and mapping to be performed automatically on each video frame. The evaluation of the mapping can be performed by computing the average difference between real locations (precisely defined by field measurements) of reference points and the projected locations of the corresponding location in the image pixel coordinates (using calibrated camera parameters).

\section{OPOS Dataset}
\label{section:dataset}
A dataset named ``{\it Objects in Public Open Spaces}'' (OPOS) is prepared based on images captured using the DRFC surveillance camera network.

\begin{figure}[bp]
\centering
\vspace{-1.5em}   
   \includegraphics[width=1\linewidth]{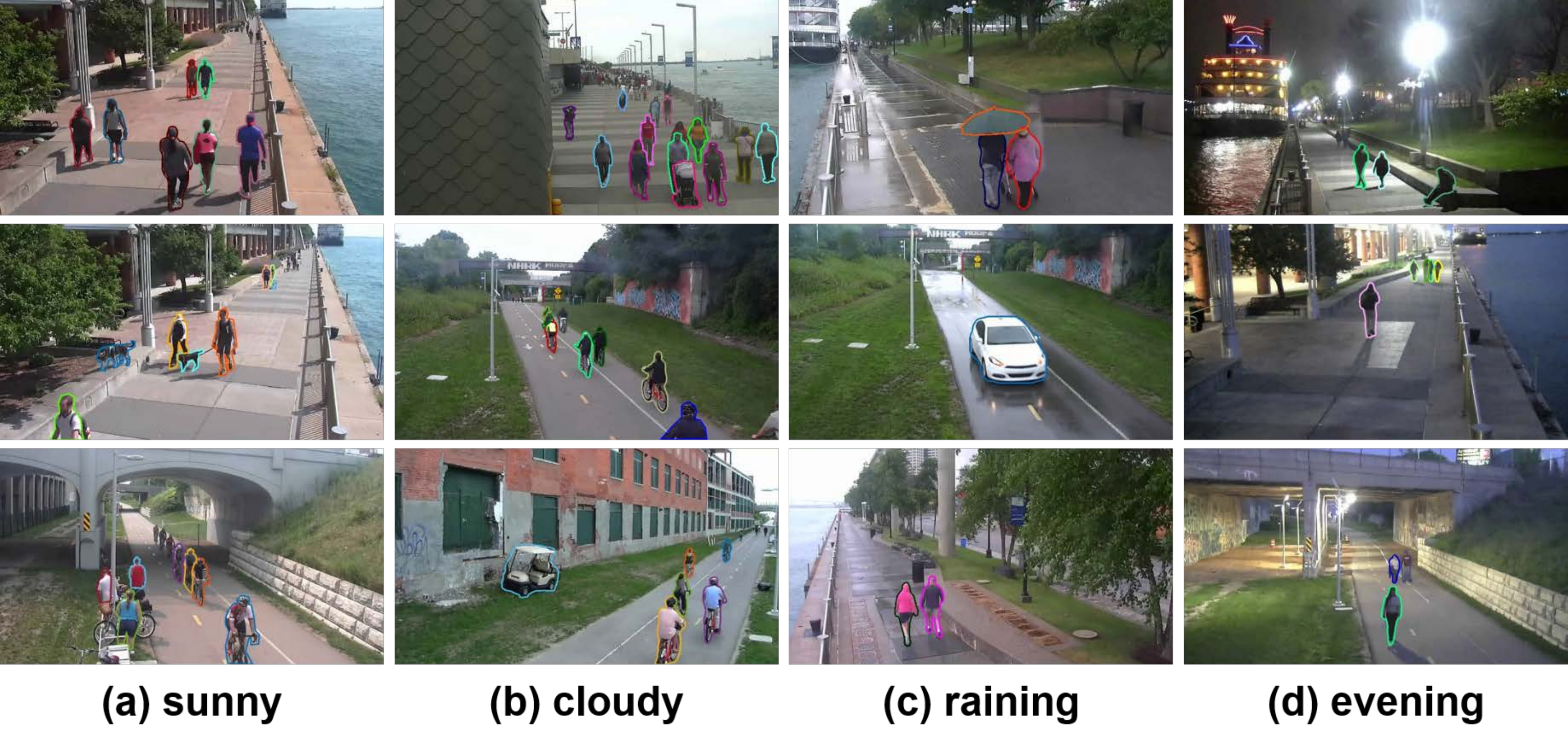}
\vspace{-1.0em}   
   \caption{Images from the DRFC surveillance cameras and the annotated objects under various illumination conditions: sunny (col 1), cloudy (col 2), rainy (col 3), and evening (col 4).}
\vspace{-1.5em}   
\label{fig:illumination_annotation}
\end{figure}

{\bf Scene selection:}  The DRFC surveillance cameras are operated at 5 or 30 fps all day (i.e. 24 hours). A total of 18 cameras (scenes) of interest are selected to collect raw video data. The video resolutions vary across different cameras (e.g. 1280$\times$720px on the Dequindre Cut, 1108$\times$832px at Cullen Plaza). Surveillance cameras capture video frames with: various traffic types (e.g. multiple pedestrians, cyclists, scooter riders), usual and rare classes (e.g. cars/trucks, skaters, dogs), different levels of camera distortion, and various illumination and weather conditions (e.g. sunny, cloudy, and rainy as shown in Fig. \ref{fig:illumination_annotation}) .

\begin{figure}[b]
\centering
\vspace{-1.2em}   
   \includegraphics[width=1\linewidth]{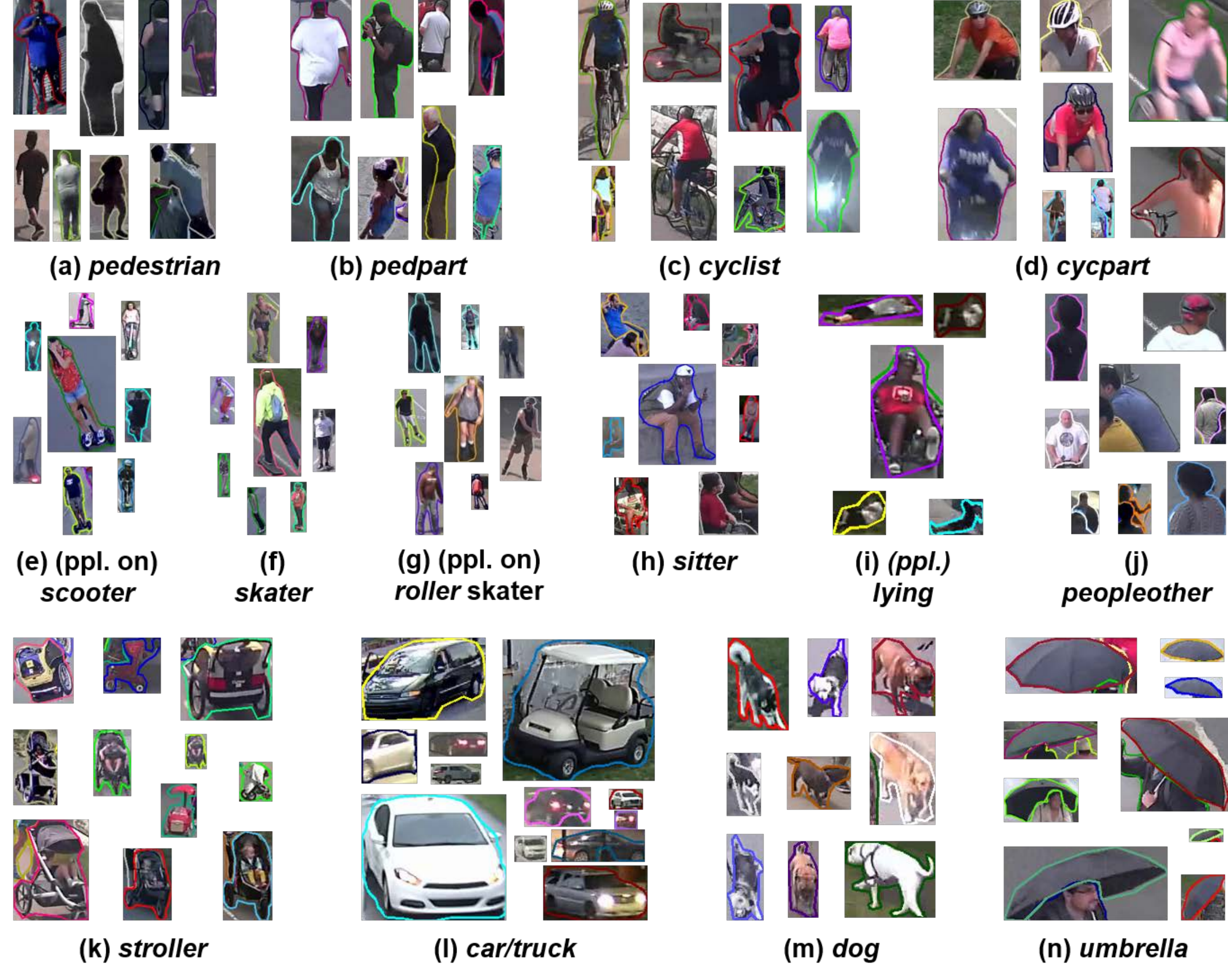}
\vspace{-1.0em}   
   \caption{Examples of different classes within the OPOS dataset.}
\label{fig:annotation_samples}
\vspace{-1.5em}   
\end{figure}

{\bf Object classes:}  The objects of interest are classified into four (4) different super-categories (i.e. {\it people}, {\it vehicle}, {\it accessory} and {\it animal}) and 15 different object classes (as shown in Fig. \ref{fig:annotation_samples}). In order to avoid the confusion for distinguishing the correct object class for each object, rules for taxonomy are defined for annotations.  For example, the definitions of the \textit{people} classes are as follows: {\it pedestrian} are people observed with crucial body parts (head, shoulder, knees, and at least one foot) connected; {\it pedpart} are people partially observed with body parts (head, shoulder, butt and knees) connected; {\it cyclist} are people riding a bicycle or with a bicycle observed with parts (head, shoulder, bicycle handlebar, and at least one bicycle wheel) connected;  {\it cycpart} are people most likely riding a bicycle partially observed with head, shoulder, crouching gesture, and bicycle handlebar/one bicycle wheel;  {\it scooterer} are people riding a scooter;  {\it skater} are people on a skateboard;  {\it roller} are people riding roller skates;  {\it sitter} are people that are sitting;  {\it people lying} are people that are lying on the grass or ground;  {\it peopleother} are people with other activities or with severe occlusions, but with head and shoulder observed.

{\bf Detector-aided annotation:}  The annotation process includes the initial collection of a seed dataset and the incremental annotation of more images by using detection models (pre-trained on the custom dataset). The seed dataset includes images on which objects of interest are first manually annotated with instance segmentations. The initial iteration of the manual annotations includes 925 images. A tight bounding box  is computed by the detected contour and a label is tagged in the form of ``super-category\_class\_obj\#'' (e.g. ``{\it person\_pedestrian\_6}''). A detector (e.g. Mask R-CNN) fine-tuned on the seed dataset is utilized to segment objects from a new batch of images. Manual revision is followed to correct any errors.   The iterations of detector-aided annotations are performed in the following steps iteratively for each 1000 additional processed images: (1) perform instance segmentation by using a newly trained Mask R-CNN detector with a threshold confidence score of 75\%; (2) filter the detection (segmentation) results by a threshold area of 600 px$^2$ and import them into LabelMe (a graphical image annotation tool) JSON file \cite{labelme2016}; (3) manually check/revise the instance segmentation (e.g. drag vertices to reshape the polygon to tightly confine a {\it pedestrian}) if needed; (4) store the new set of data and use it to train a new detection model. 

{\bf Data format and privacy protection:} Labels and annotations are stored in the COCO object detection (segmentation) format, including 4 sections (i.e. info, licenses, images, annotations). Each annotation section includes information of object id, category id, segmentation, bounding box, etc. To ensure the privacy of the dataset, the cascaded convolutional network \cite{zhang2016face} is used to detect human faces within the dataset and the results are double checked manually to cover human faces. The final results are used to blur human faces in the dataset to preserve user privacy.

{\bf Statistical analysis:} The statistics of the annotations in the OPOS dataset are analyzed and presented in this section. The dataset 
(Table \ref{tab:num_allobjects}) includes various people, vehicles, and other mobile object classes. The total number of segmented objects is 18.9K and the most frequent objects are {\it pedestrian} and {\it cyclist}.  The annotated objects are under various weather conditions (e.g.  6.7\%  rainy, 45.8\% cloudy, and 47.4\% sunny) during spring and summer seasons.  Also, 14.1\%  of the annotations are during the evening with street lights on, and 7.2\% of the annotations are small-size objects.

\begin{table}[tp]
\centering
\vspace{-1.5em}   
\footnotesize
\begin{tabular}{p{1.8cm}p{1cm}p{0.7cm}p{0.7cm}p{0.6cm}p{0.6cm}}
\toprule
\multirow{2}{*}{Object Class} & \multirow{2}{*}{\# Obj.} & \multicolumn{2}{c}{Area (px$^2$)} & \multicolumn{2}{c}{Aspect ratio} \\ \cmidrule(l{10pt}r{10pt}){3-4} \cmidrule(l{10pt}r{10pt}){5-6}
                              &                         & \textit{mean}  & \textit{std}  & \textit{mean}   & \textit{std}   \\  \midrule
\textit{pedestrian}           & 9675                    & 4659           & 4003          & 2.81            & 0.48           \\
\textit{cyclist}              & 5092                    & 6432           & 6505          & 2.56            & 0.42           \\
\textit{scooter}              & 466                     & 5056           & 5231          & 2.83            & 0.51           \\
\textit{skater}               & 43                      & 5746           & 4043          & 2.99            & 0.53           \\
\textit{sitter}               & 704                     & 3719           & 2844          & 1.52            & 0.44           \\
\textit{peopleother}          & 736                     & 6449           & 6433          & 1.45            & 0.55           \\
\textit{pedpart}              & 487                     & 8475           & 7063          & 2.25            & 0.54           \\
\textit{cycpart}              & 192                     & 8321           & 7902          & 1.50            & 0.56           \\
\textit{roller}               & 86                      & 4716           & 4119          & 2.55            & 0.43           \\
\textit{lay}                  & 6                       & 2344           & 1963          & 0.80            & 0.51           \\
\textit{stroller}             & 216                     & 4955           & 3848          & 1.64            & 0.47           \\
\textit{car}                  & 736                     & 12861          & 14521         & 0.80            & 0.24           \\
\textit{vehicleother}         & 51                      & 6000           & 6319          & 1.24            & 0.34           \\
\textit{dog}                  & 303                     & 1543           & 1070          & 1.60            & 0.57           \\
\textit{umbrella}             & 109                     & 4449           & 5257          & 0.48            & 0.16          \\ \bottomrule
\end{tabular}
\caption{Numbers of annotations per object class in OPOS.}
\label{tab:num_allobjects}
\vspace{-2em}   
\end{table}
\normalsize

In real situations, surveillance cameras can capture occluded objects or truncated ones at the image boundary (due to objects moving in or out of view).  Sometimes, even with occlusions or truncation, humans can still infer the object as a partially viewed object (e.g. {\it pedestrian} or {\it cyclist}) by referring to some visual cues (e.g. the gesture of standing up straight or the gesture of holding the steering bar of a bicycle). Hence, truncated/occluded people classes (e.g. denoted as {\it pedpart} and {\it cycpart}) are also included in the dataset to check whether a detection model can distinguish between visual cues. 

\section{Experiments} \label{section:exp}
\subsection{Dataset Setup for Training} \label{section:exp_dataset}
The OPOS custom dataset includes 7826 images. Pre-training on public datasets is preferred before performing transfer learning on the OPOS dataset. The weights of the ResNet backbones are first pre-trained on the ImageNet-1k dataset.  The weights of most of the models (except the first model in Table \ref{tab:pretrain_dataset}) are additionally pre-trained on the coco\_2017\_train dataset (pretrained weights are from the Detectron website \cite{Detectron2018}). The training and test sets of the OPOS dataset are split by the ratio of 9:1. 

{\bf Fine-tuning schedule:}  the pre-trained weights are fine-tuned on the OPOS dataset using the maskrcnn-benchmark platform \cite{massa2018mrcnn} with a NVIDIA 1070 GPU.  The mini batch size is set as 2 images/batch,  and horizontal flipping is adopted for data augmentation. The schedule including 90K iterations starts at a learning rate of 0.0025, decreases by a factor of 0.1 after both 60K and 80K iterations, and finally terminates at 90k iterations.  This schedule results in 25.56 epochs of training on the training data set.  The ResNet backbone has five stages \cite{he2016deep} (denoted as conv1-5) with each stage consisting of several stacked residual blocks. The first two stages (i.e. conv1 and conv2) have been frozen, while the weights of the other stages are kept flexible for updating. 

\begin{table*}[tp]
\centering
\footnotesize
\vspace{-2.0em}   
\begin{tabular}{@{}p{2.4cm}p{0.8cm}p{.8cm}p{.8cm}p{.8cm}p{.8cm}p{.8cm}p{.8cm}p{.8cm}p{.8cm}p{.8cm}p{.8cm}@{}}
\toprule 
                                                                                 &                                                                            & \multicolumn{8}{c}{mAP$_c$  per ppl. class}                                                                                                                                                                                                                                                                                    &                                                                             &                                                                            \\ \cmidrule(l{10pt}r{10pt}){3-10}
\multirow{-2}{*}{\begin{tabular}[c]{@{}l@{}}Treatment of \\  Occl./Trun. Classes\end{tabular}}   & \multirow{-2}{*}{\begin{tabular}[c]{@{}c@{}}bbox\\     /segm\end{tabular}} & \textit{ped.}                           & \textit{cycl.}                          & \textit{scoot.}                         & \textit{skater}                          & \textit{sitter}                         & \textit{other}                          & \textit{pedpart}               & \textit{cycpart}               & \multirow{-2}{*}{\begin{tabular}[c]{@{}c@{}}mAP \\     (ppl.)\end{tabular}} & \multirow{-2}{*}{\begin{tabular}[c]{@{}c@{}}mAP \\ (overall)\end{tabular}} \\
\midrule
                                                                                 & \textit{bbox}                                                              & 77.5\%                                  & 81.6\%                                  & 74.7\%                                  & \textbf{63.3\%}                         & \textbf{65.7\%}                         & 53.4\%                                  & NA                             & NA                             & 69.4\%                                                                      & 67.7\%                                                                     \\
\multirow{-2}{*}{\begin{tabular}[c]{@{}l@{}}Merging (part to \\ whole)\end{tabular}}       & \cellcolor[HTML]{EFEFEF}\textit{segm}                                      & \cellcolor[HTML]{EFEFEF}74.7\%          & \cellcolor[HTML]{EFEFEF}81.2\%          & \cellcolor[HTML]{EFEFEF}74.3\%          & \cellcolor[HTML]{EFEFEF}\textbf{66.4\%} & \cellcolor[HTML]{EFEFEF}\textbf{59.4\%} & \cellcolor[HTML]{EFEFEF}53.9\%          & \cellcolor[HTML]{EFEFEF}NA     & \cellcolor[HTML]{EFEFEF}NA     & \cellcolor[HTML]{EFEFEF}68.3\%                                              & \cellcolor[HTML]{EFEFEF}66.2\%                                             \\
                                                                                 & \textit{bbox}                                                              & \textbf{78.5\%}                         & \textbf{81.6\%}                         & \textbf{81.4\%}                         & 57.4\%                                  & 64.7\%                                  & \textbf{61.5\%}                         & NA                             & NA                             & \textbf{70.9\%}                                                             & \textbf{68.4\%}                                                            \\
\multirow{-2}{*}{\begin{tabular}[c]{@{}l@{}}Filtering (part to \\ pplother)\end{tabular}} & \cellcolor[HTML]{EFEFEF}\textit{segm}                                      & \cellcolor[HTML]{EFEFEF}\textbf{75.4\%} & \cellcolor[HTML]{EFEFEF}\textbf{81.8\%} & \cellcolor[HTML]{EFEFEF}\textbf{79.7\%} & \cellcolor[HTML]{EFEFEF}65.1\%          & \cellcolor[HTML]{EFEFEF}59.4\%          & \cellcolor[HTML]{EFEFEF}\textbf{61.3\%} & \cellcolor[HTML]{EFEFEF}NA     & \cellcolor[HTML]{EFEFEF}NA     & \cellcolor[HTML]{EFEFEF}\textbf{70.5\%}                                     & \cellcolor[HTML]{EFEFEF}\textbf{67.0\%}                                    \\
                                                                                 & \textit{bbox}                                                              & 78.3\%                                  & 81.4\%                                  & 77.4\%                                  & 53.6\%                                  & 64.9\%                                  & 56.1\%                                  & 58.4\%                         & 38.0\%                         & 63.5\%                                                                      & 64.6\%                                                                     \\
\multirow{-2}{*}{\begin{tabular}[c]{@{}l@{}} Separating (keep\\ part  classes)\end{tabular}}  & \cellcolor[HTML]{EFEFEF}\textit{segm}                                      & \cellcolor[HTML]{EFEFEF}75.2\%          & \cellcolor[HTML]{EFEFEF}81.6\%          & \cellcolor[HTML]{EFEFEF}76.6\%          & \cellcolor[HTML]{EFEFEF}50.5\%          & \cellcolor[HTML]{EFEFEF}59.3\%          & \cellcolor[HTML]{EFEFEF}55.5\%          & \cellcolor[HTML]{EFEFEF}59.0\% & \cellcolor[HTML]{EFEFEF}37.6\% & \cellcolor[HTML]{EFEFEF}61.9\%                                              & \cellcolor[HTML]{EFEFEF}62.7\% \\
\bottomrule                                            
\end{tabular}
\caption{Comparison of models with the same backbone (ResNet50-FPN) trained on the same datasets (Imagenet+COCO+OPOS) with different treatments of occluded/truncated classes.}
\label{tab:truncation}
\vspace{-1.8em}   
\end{table*}
\normalsize

\subsection{Study of Detection Models} \label{section:exp_models}

{\bf Treatment of similar people classes:}  There are 10 classes within the {\it people} super-category.  However, there are many similarities across closely related classes within the same super-category; for example, a {\it roller} (people wearing roller skates) is very similar to a {\it pedestrian} except slight differences on the foot wear.  If they are treated as two separate classes for training, the detection results are expected to be inconsistent. For example, a {\it roller} would be detected incorrectly as a {\it pedestrian} (most likely to happen) in far field, but once the person is approaching the camera, the person would be identified as {\it roller} once the model spotted the nuances in foot wear.  This would cause trouble in practice for user detection or future tracking tasks.  Hence, in the current stage of research, {\it roller} is merged into {\it pedestrian}.

\begin{table}[bp]
\footnotesize
\centering
\vspace{-1.5em}   
\begin{tabular}{p{0.8cm}p{0.5cm}p{0.45cm}p{0.45cm}p{0.45cm}p{0.45cm}p{0.45cm}p{0.45cm}p{0.45cm}}
\toprule
                                                                                   &                                                                            & \multicolumn{6}{c}{mAP$_c$  per ppl. class}                                                                                                                                                            &                                \\ \cmidrule(l{10pt}r{10pt}){3-8}
\multirow{-2}{*}{Datasets}                                                          & \multirow{-2}{*}{\begin{tabular}[c]{@{}c@{}}bbox\\     /segm\end{tabular}} & \textit{ped.}                  & \textit{cycl.}                 & \textit{scoot.}                & \textit{skat.}                 & \textit{sitter}                & \textit{other}                 &\multirow{-2}{*}{\begin{tabular}[c]{@{}c@{}}mAP \\ (ppl.)\end{tabular}}   \\  \midrule
                                                                                   & \textit{bbox}                                                              & 76.1\%                         & 80.3\%                         & 76.3\%                         & 48.5\%                         & 65.3\%                         & 49.3\%                         & 66.0\%                         \\
\multirow{-2}{*}{\begin{tabular}[c]{@{}l@{}}imagenet+\\     opos\end{tabular}}      & \cellcolor[HTML]{EFEFEF}\textit{segm}                                      & \cellcolor[HTML]{EFEFEF}73.3\% & \cellcolor[HTML]{EFEFEF}80.4\% & \cellcolor[HTML]{EFEFEF}78.3\% & \cellcolor[HTML]{EFEFEF}44.3\% & \cellcolor[HTML]{EFEFEF}56.4\% & \cellcolor[HTML]{EFEFEF}49.4\% & \cellcolor[HTML]{EFEFEF}63.7\% \\
                                                                                   & \textit{bbox}                                                              & 77.5\%                         & 81.6\%                         & 74.7\%                         & 63.3\%                         & 65.7\%                         & 53.4\%                         & \textbf{69.4\%}                         \\
\multirow{-2}{*}{\begin{tabular}[c]{@{}l@{}}imagenet+\\     coco+opos\end{tabular}} & \cellcolor[HTML]{EFEFEF}\textit{segm}                                      & \cellcolor[HTML]{EFEFEF}74.7\% & \cellcolor[HTML]{EFEFEF}81.2\% & \cellcolor[HTML]{EFEFEF}74.3\% & \cellcolor[HTML]{EFEFEF}66.4\% & \cellcolor[HTML]{EFEFEF}59.4\% & \cellcolor[HTML]{EFEFEF}53.9\% & \cellcolor[HTML]{EFEFEF}\textbf{68.3\%}  \\
\bottomrule
\end{tabular}
\caption{Comparison of models with the same backbone (ResNet50-FPN) trained on different datasets.}
\label{tab:pretrain_dataset}
\vspace{-2.2em}   
\end{table}
\normalsize

{\bf Treatments of occluded/truncated classes:} The occluded/truncated classes (i.e. {\it pedpart} and {\it cycpart}, denoted as part classes) consist of 3.6\% of the overall objects.  There is a need to study the influence of different treatments of the part classes.  Three arrangements are made here: (1) ``merging" -- treating the part classes as the corresponding whole classes (i.e. {\it pedpart} is treated as {\it pedestrian}, and {\it cycpart} is treated as {\it cyclist}); (2) ``filtering" -- treat the part classes as {\it peopleother} class (i.e. both {\it pedpart} and {\it cycpart} are treated as {\it peopleother}); (3) ``separating" -- treating the part classes as individual classes (i.e. {\it pedpart} and {\it cycpart} classes are kept as separate classes).  As shown in Table \ref{tab:truncation}, the detection performances are very close for merging and filtering treatments, and the two outperform the separating treatment by more than 3\% in mAP.  This might be due to the strong similarity between the truncated classes and the corresponding whole classes.  The results show that the treatment of either filtering or merging reduces the confusion in detection for these classes, and thus improves the detection performance.  Although the merging treatment slightly underperforms the filtering treatment (e.g. 1.3\% lower in bbox mAP and 0.8\% lower in segmentation mAP), the merging treatment is more logical in understanding and practical usage.  Hence, the rest of the study will be performed using the merging treatment of the part classes.

{\bf Effect of pre-training:} Some of the models are directly fine-tuned on the OPOS dataset by using the pre-trained weights on the ImageNet dataset (denoted as “ImageNet+OPOS”) and the other models use pre-trained weights on both ImageNet and coco\_2017\_train datasets (denoted as “ImageNet+COCO+OPOS”).  As shown in Table \ref{tab:pretrain_dataset}, the APs for various people (ppl.) classes observe a general increase with the additional pre-training on the coco\_2017\_train dataset.  The increases for mAP of the overall people classes are 3.4\% in bbox detection and 4.6\% in segmentation detection, respectively.  Hence, pre-training on more datasets would benefit the ultimate performance of the detection models.

{\bf Impact of backbone architectures:}  The detection models of Mask R-CNN with different backbone architectures are trained on ImageNet+COCO+OPOS, as shown in Table \ref{tab:backbone}.  The results show that ResNet50-FPN (``-FPN" denotes a head from the second half of FPN) outperforms RestNet50-C4 (``-C4" denotes a head from the fourth module of ResNet50) with an increase of 2.9\% in bbox detection and 4.6\% in segmentation detection for overall classes.  The results show the advantages of using combined feature maps at different scales (from FPN head) over using a single feature map (from C4 head).  Backbone architectures with various depths (e.g. ResNet50-FPN and ResNet101-FPN) are also compared (where ``50'' and ``101'' refer to the numbers of convolutional layers).  ResNet50-FPN outperforms ResNet101-FPN with the same training protocol (90k iterations). A prolonged training schedule (180k iterations) improves ResNet101-FPN by 4.3\% in bbox detection.   

\begin{table}[bp] 
\footnotesize
\centering
\vspace{-1.8em}   
\begin{tabular}{p{1.8cm}p{0.75cm}p{0.8cm}p{0.8cm}p{0.8cm}p{0.85cm}}  

\toprule
                                                                                    &                                                                            & \multicolumn{4}{c}{mAP$^{size}$ for overall cls.}                                                                                                                                   \\ \cmidrule(l{10pt}r{10pt}){3-6}
\multirow{-2}{*}{\begin{tabular}[c]{@{}l@{}}Backbone\\   architecture\end{tabular}} & \multirow{-2}{*}{\begin{tabular}[c]{@{}c@{}}bbox\\     /segm\end{tabular}} & small                          & medium                         & large                          & all-size                            \\ \midrule
                                                                                    & \textit{bbox}                                                              & 50.4\%                         & 66.4\%                         & 70.9\%                                  & 64.8\%                                  \\
\multirow{-2}{*}{ResNet50-C4}                                                       & \cellcolor[HTML]{EFEFEF}\textit{segm}                                      & \cellcolor[HTML]{EFEFEF} 48.0\% & \cellcolor[HTML]{EFEFEF} 62.0\% & \cellcolor[HTML]{EFEFEF}77.3\%          & \cellcolor[HTML]{EFEFEF}61.6\%          \\
                                                                                    & \textit{bbox}                                                              & \textbf{57.6\%}                                  & 67.8\%                                  & \textbf{77.4\%}                         & 67.7\%                     \\
\multirow{-2}{*}{ResNet50-FPN}                                                      & \cellcolor[HTML]{EFEFEF}\textit{segm}                                      & \cellcolor[HTML]{EFEFEF}\textbf{50.2\%}          & \cellcolor[HTML]{EFEFEF}66.3\%          & \cellcolor[HTML]{EFEFEF}\textbf{81.6\%} & \cellcolor[HTML]{EFEFEF}\textbf{66.2\%} \\
                                                                                    & \textit{bbox}                                                              & 53.8\%                                  & 66.0\%                                  & 73.8\%                                  & 63.8\%                                  \\
\multirow{-2}{*}{ResNet101-FPN}                                                     & \cellcolor[HTML]{EFEFEF}\textit{segm}                                      & \cellcolor[HTML]{EFEFEF}49.4\%          & \cellcolor[HTML]{EFEFEF}63.6\%          & \cellcolor[HTML]{EFEFEF}74.0\%          & \cellcolor[HTML]{EFEFEF}61.8\%         \\
                                                                                       & \textit{bbox}                         & 53.3\%                         & \textbf{70.0}\%                         & 76.0\%                         & \textbf{68.1\%} \\
\multirow{-2}{*}{\begin{tabular}[c]{@{}l@{}}ResNet101-FPN\\   (2x train)\end{tabular}} & \cellcolor[HTML]{EFEFEF}\textit{segm} & \cellcolor[HTML]{EFEFEF}49.8\% & \cellcolor[HTML]{EFEFEF}\textbf{66.8\%} & \cellcolor[HTML]{EFEFEF}77.2\% & \cellcolor[HTML]{EFEFEF} 65.5\%        \\ 

\bottomrule
\end{tabular}
\caption{Comparison of Mask R-CNN models with different backbones (trained on ImageNet+COCO+OPOS).}
\label{tab:backbone}
\vspace{-2.2em}   
\end{table}
\normalsize

\subsection{Error Diagnosis of the Detection Model} \label{section:exp_error}

\begin{figure} [bp]
        \vspace{-1.8em}   
        \centering
        
        \begin{subfigure}[b]{0.22\textwidth}
            \centering
            \includegraphics[width=\textwidth]{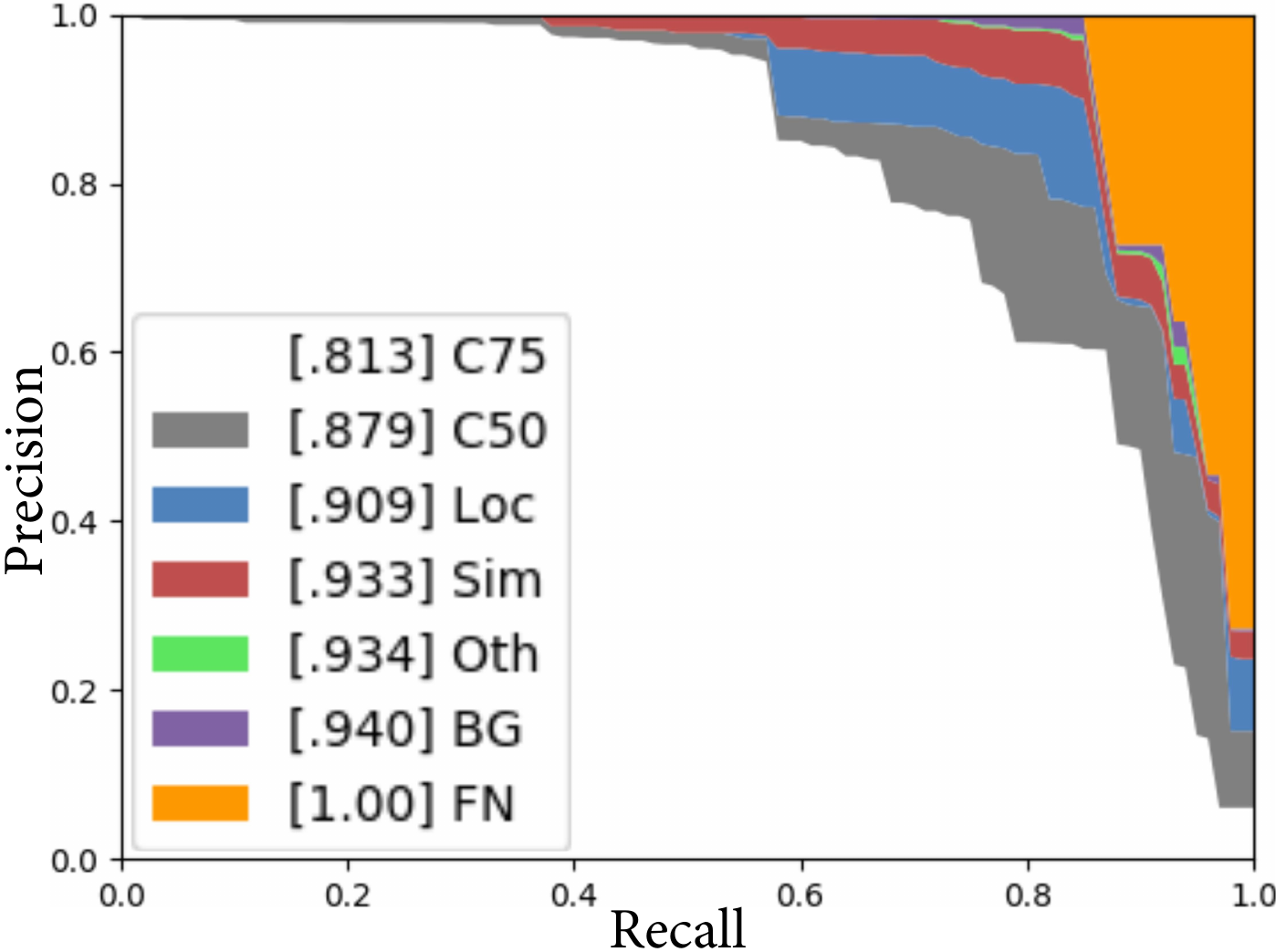}
            \vspace{-1.8em}   
            \caption[]%
            {{\footnotesize overall-all}}    
            \label{fig:PR_overall}
        \end{subfigure}
        \quad
        \begin{subfigure}[b]{0.22\textwidth}  
            \centering 
            \includegraphics[width=\textwidth]{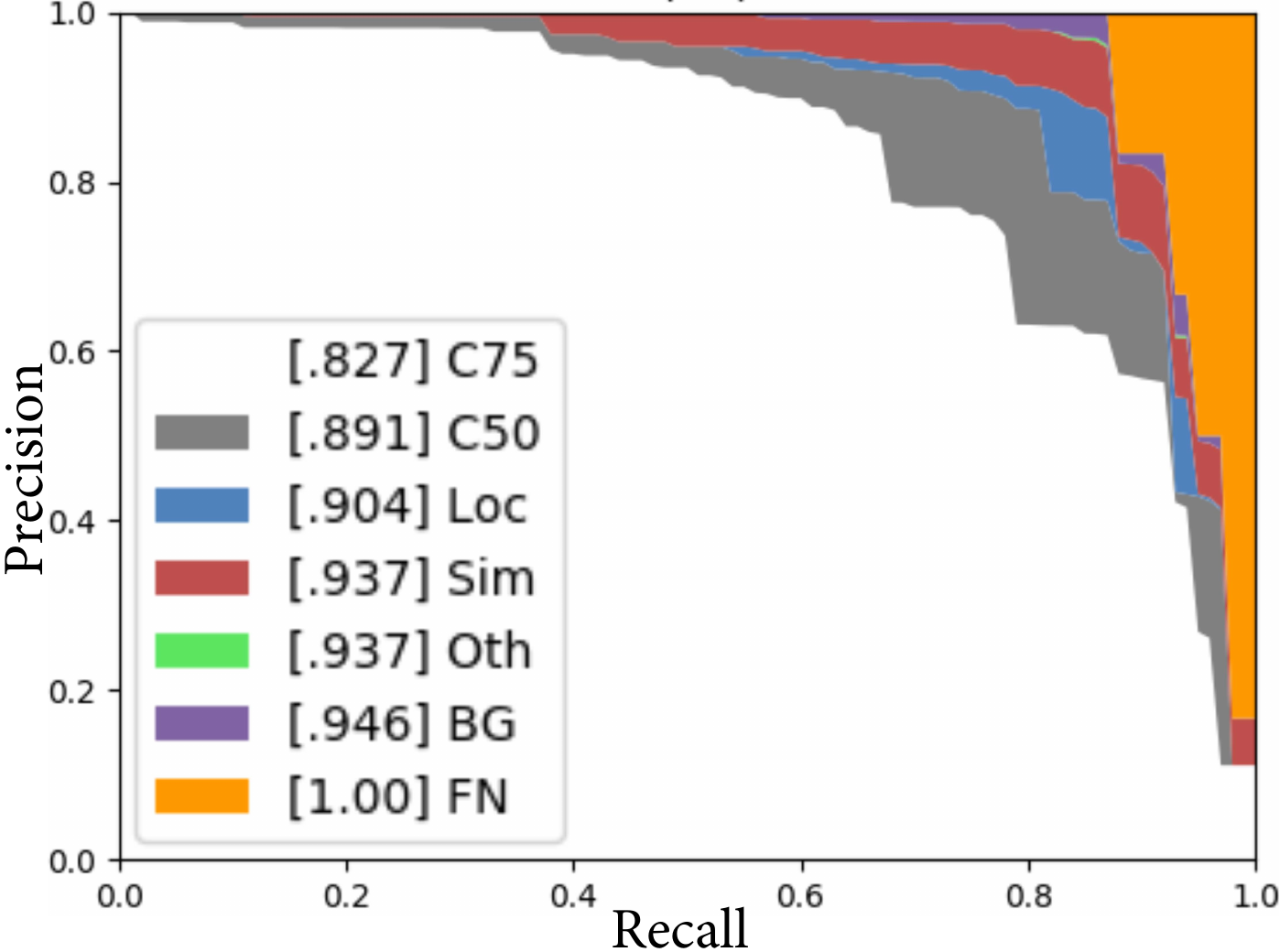}
            \vspace{-1.8em}   
            \caption[]%
            {{\footnotesize {\it people}-all}}    
            \label{fig:PR_people}
        \end{subfigure} \\
        
        \begin{subfigure}[b]{0.22\textwidth}
            \centering
            \includegraphics[width=\textwidth]{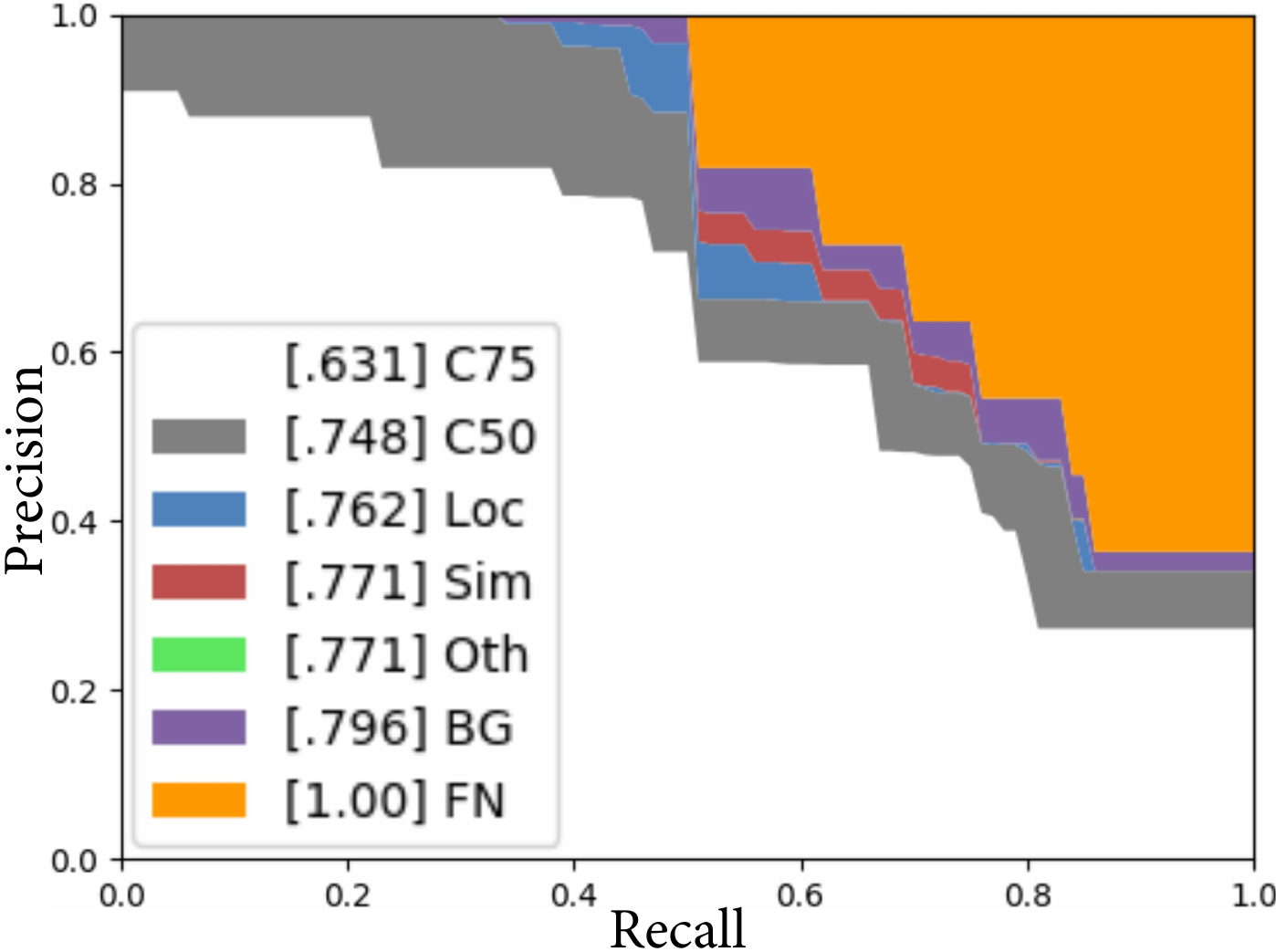}
            \vspace{-1.8em}   
            \caption[]%
            {{\footnotesize overall-small}}    
            \label{fig:PR_small}
        \end{subfigure}
        \quad
        \begin{subfigure}[b]{0.22\textwidth}  
            \centering 
            \includegraphics[width=\textwidth]{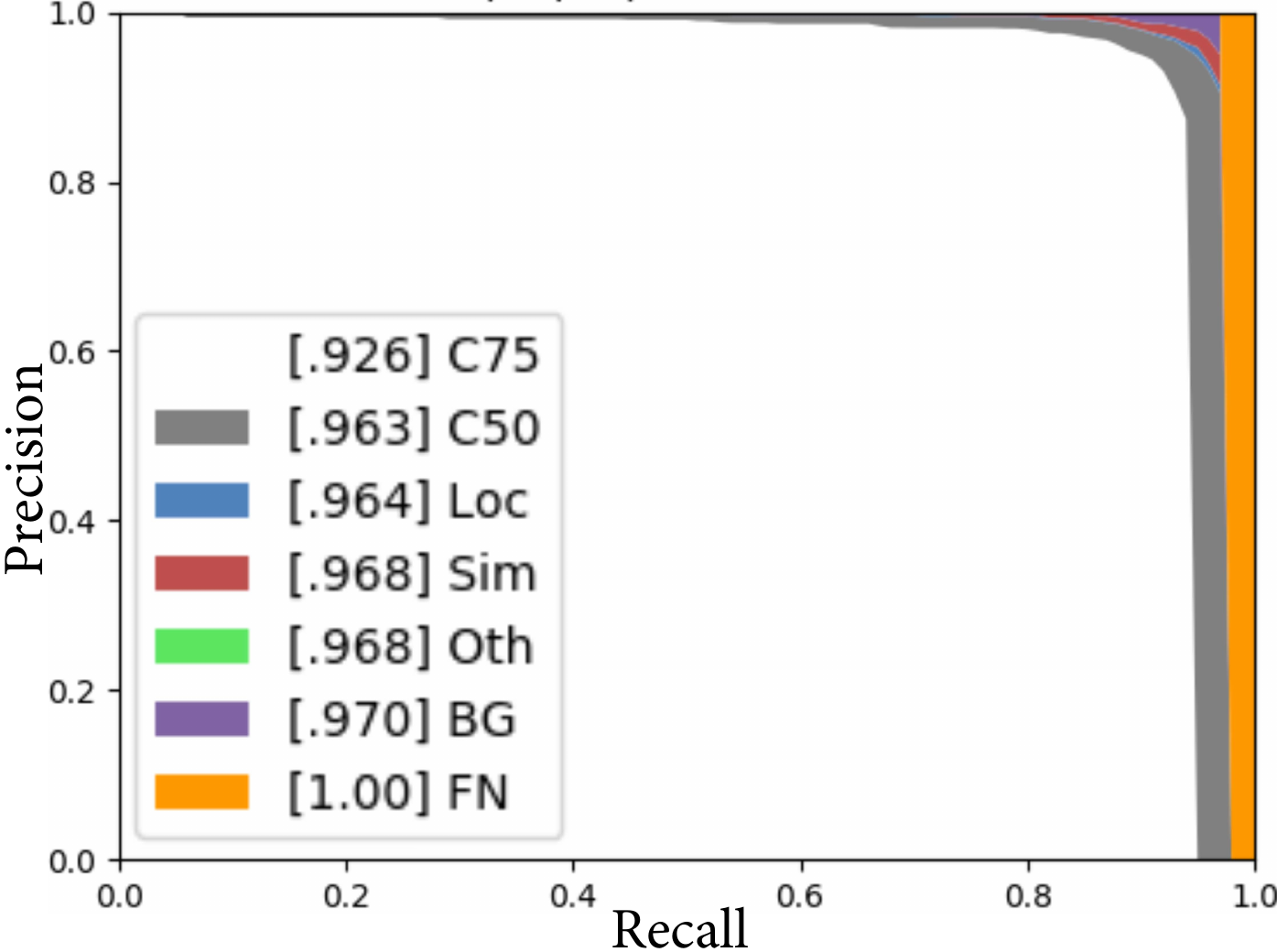}
            \vspace{-1.8em}   
            \caption[]%
            {{\footnotesize {\it pedestrian}-all}}    
            \label{fig:PR_pedestrian}
        \end{subfigure} \\
   
        \begin{subfigure}[b]{0.22\textwidth}   
            \centering 
            \includegraphics[width=\textwidth]{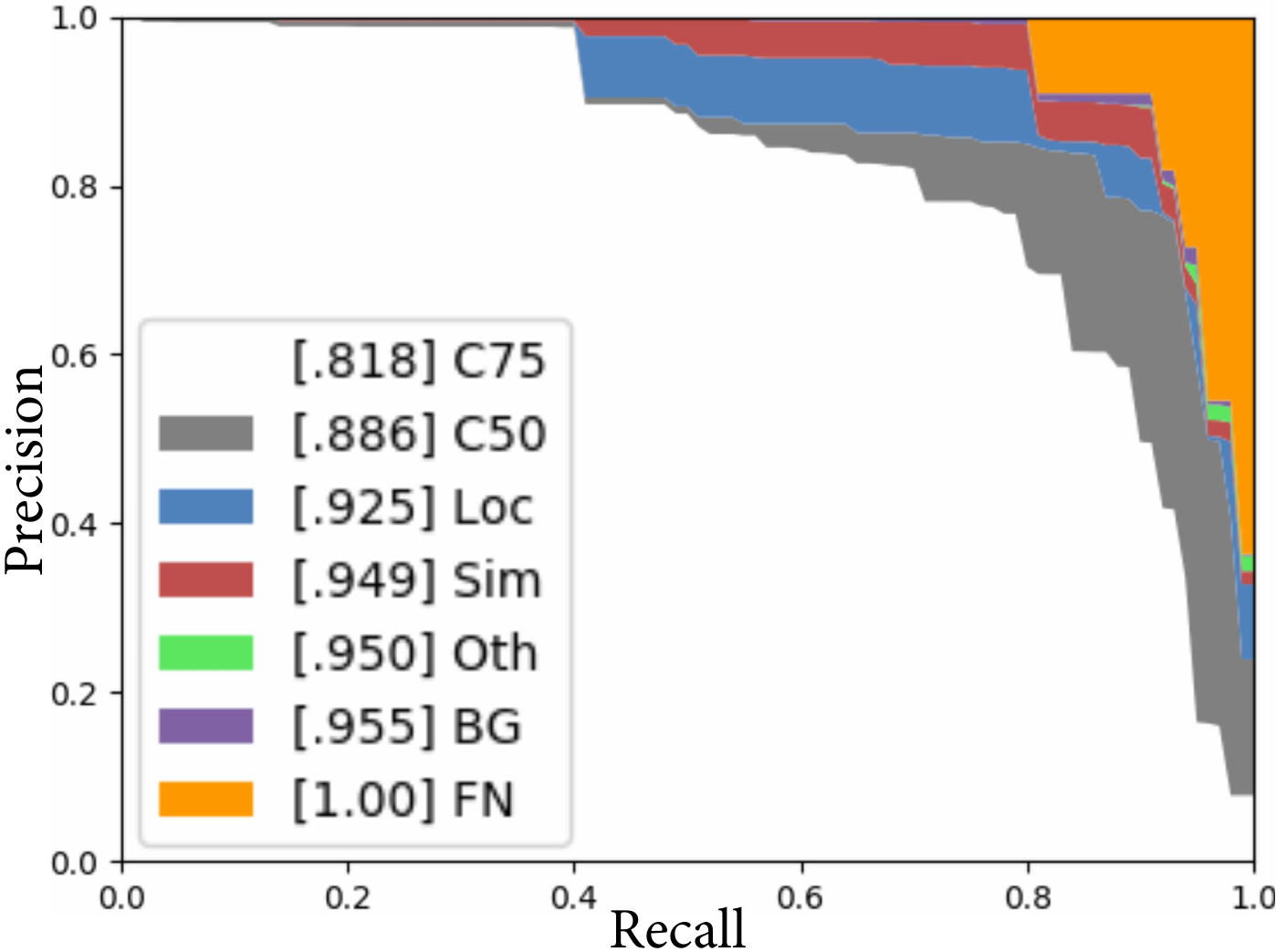}
            \vspace{-1.8em}   
            \caption[]%
            {{\footnotesize overall-medium}}    
            \label{fig:PR_medium}
        \end{subfigure}
        \quad
        \begin{subfigure}[b]{0.22\textwidth}   
            \centering 
            \includegraphics[width=\textwidth]{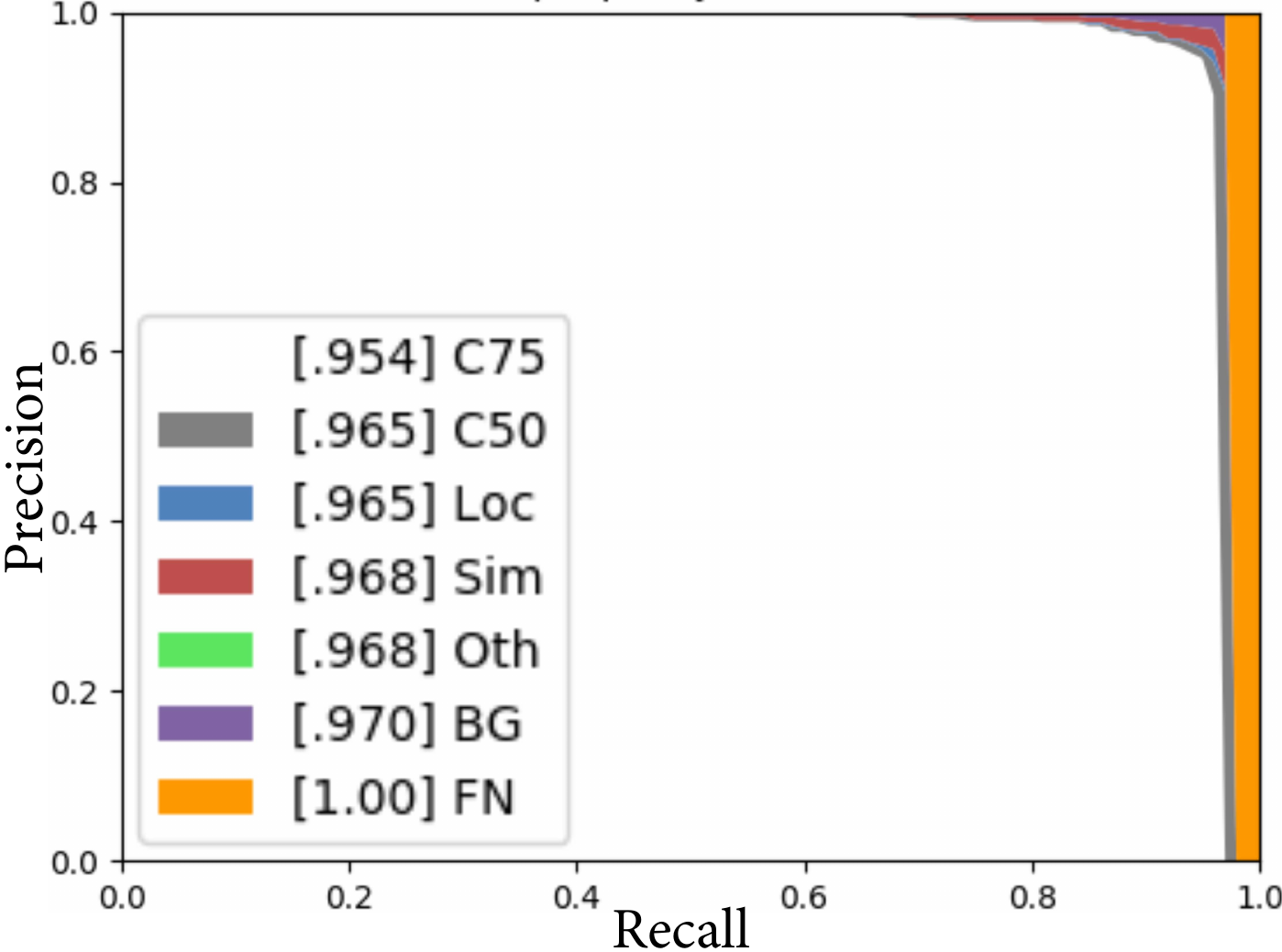}
            \vspace{-1.8em}   
            \caption[]%
            {{\footnotesize {\it cyclist}-all}}    
            \label{fig:PR_cyclist}
        \end{subfigure}

        \begin{subfigure}[b]{0.22\textwidth}
            \centering
            \includegraphics[width=\textwidth]{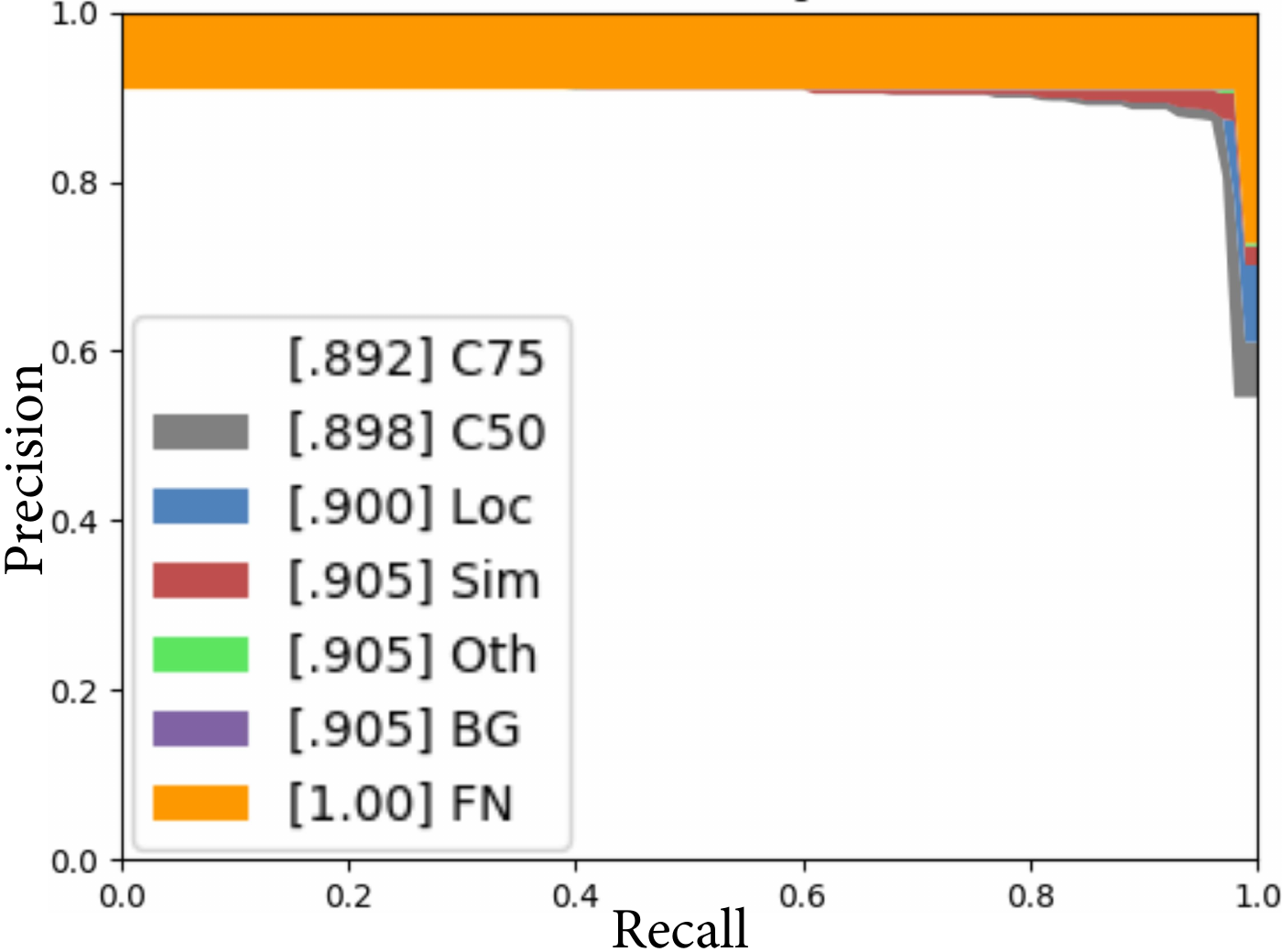}
            \vspace{-1.8em}   
            \caption[]%
            {{\footnotesize overall-large}}    
            \label{fig:PR_large}
        \end{subfigure}
        \quad
        \begin{subfigure}[b]{0.22\textwidth}  
            \centering 
            \includegraphics[width=\textwidth]{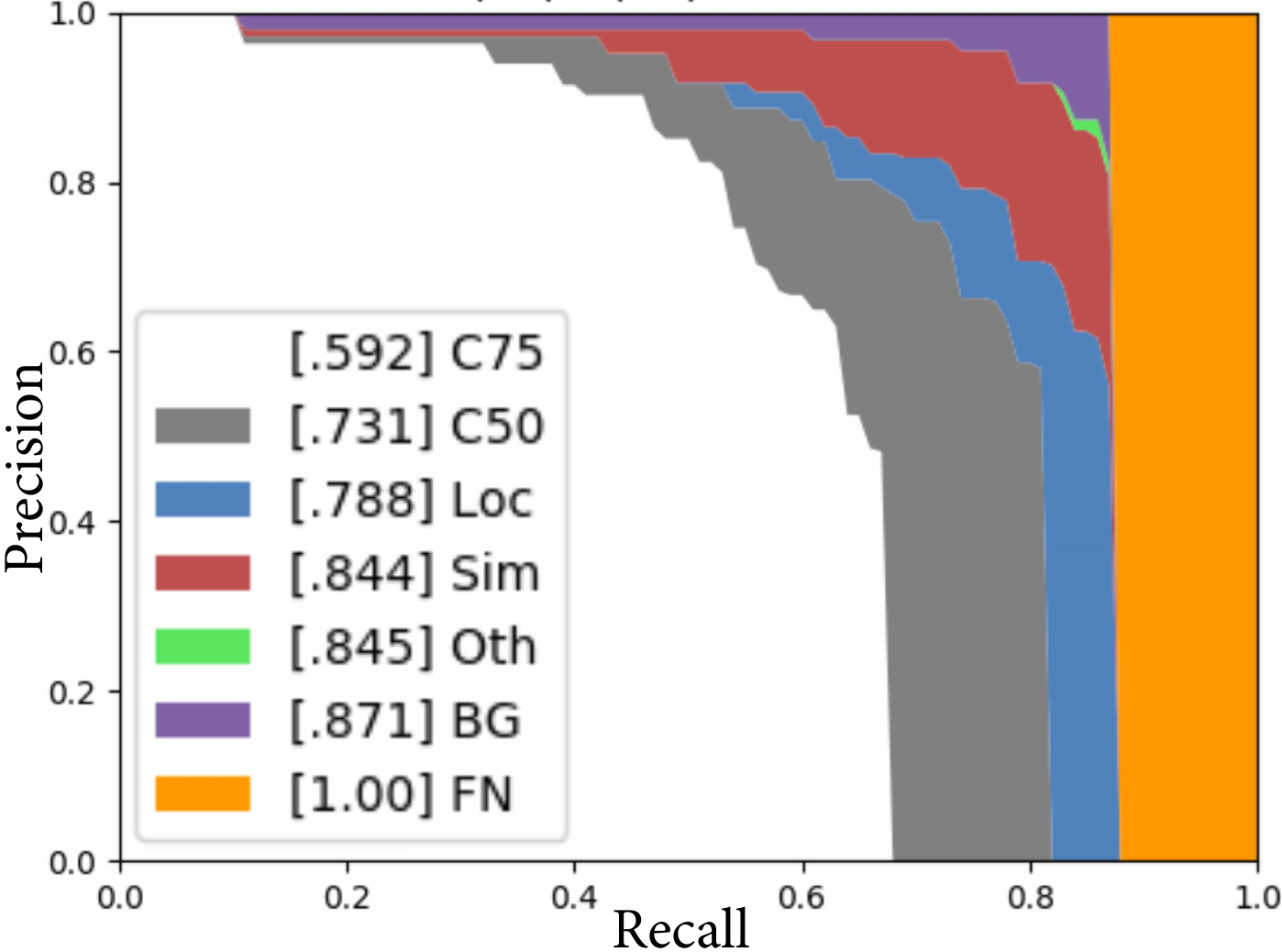}
            \vspace{-1.8em}   
            \caption[]%
            {{\footnotesize {\it peopleother}-all}}    
            \label{fig:PR_peopleother}
        \end{subfigure} \\

        \caption{Precision-recall curves of (left col.) overall classes at different sizes, and (right col.) specific people classes at all sizes.} 
        \label{fig:PR_all}
        \vspace{-2.2em}   
    \end{figure}

The evaluation of a trained baseline model of Mask R-CNN is demonstrated here.  The detection task requires detecting 11 object classes (6 classes of people with part classes merged into whole classes) which are a subset of all 15 classes in the OPOS dataset. The trained detector is evaluated on the test dataset consisting of 783 images. If not stated specifically, AP is referred as bbox AP in the rest of the study.  The mAP metrics for each people class, overall people classes, and overall classes are shown in Table \ref{tab:truncation}. PR curves are used to study the errors for the detection model in detail, as presented in Fig. \ref{fig:PR_all} \cite{hoiem2012diagnosing}. The area under each PR curve represents the AP under a specific criteria.  ``C75" represents the PR curve at IoU=0.75; ``C50" represents the PR curve at IoU=0.50; ``Loc" represents the PR curve at IoU=0.10, when localization error is ignored (denoted as ``perfect localization''), but not duplicate detection; ``Sim" represents the PR (IoU=0.10 for ``Sim", ``Oth", ``BG", and ``FN") after super-category false positives (class confusion within same super-category) are removed; ``Oth" represents the PR after all intra-class false positives (both within super-categories and cross super-categories, denoted as ``class confusion'') are removed; ``BG" represents the PR after all background false positives are additionally removed; ``FN" represents the PR after false negatives are additionally removed. Among all the PR plots, results of the overall object classes and \textit{people} classes are of the most interest to evaluate the proposed detection model.

{\bf Detection performance for overall objects:} The overall AP$^{0.75}$ of the trained Mask R-CNN detector is 81.3\% and perfect localization would improve AP to 90.9\%.  While the effects of class confusion (2.5\% within super-categories) and background confusion (0.6\%) are very trivial compared to the effect of perfect localization.  In general, the APs for small objects (e.g. 63.1\% in AP$^{0.75}$) are poorer than those for both medium (e.g. 81.8\% in AP$^{0.75}$) and large (e.g. 89.2\% in AP$^{0.75}$) objects.  For small object detection (Fig. \ref{fig:PR_small}), there is a considerable amount of localization errors (13.1\%) and false negatives (20.4\%).  Future improvement of the detection model could focus on reducing localization error and false negatives. For example, more small object annotations could be added to OPOS.  However, detecting small objects intrinsically is a hard problem, because small objects include much less information (pixel-wise) and sometimes cannot provide necessary distinctive features for class detection.  For medium and large object detection, the overall performances are satisfactory with plum shaped PR curves and high APs.  The potential improvement could be targeted at reducing localization (LOC) error for medium objects (currently 10.7\% LOC error) and reducing the false negative (FN) predictions for large objects (currently 9.5\% FN error). 

{\bf Detection performance for people:}  No substantial drawback is observed for people detection in general.  The detection performance for some common people classes, for example, newly merged {\it pedestrian} class (including {\it roller} and {\it pedpart}), newly merged {\it cyclist} class (including {\it cycpart}), and {\it peopleother} class can be seen in Fig. \ref{fig:PR_pedestrian}, \ref{fig:PR_cyclist}, and \ref{fig:PR_peopleother}, respectively.  The precision for predicting {\it pedestrian} and {\it cyclist} classes are high with AP$^{0.75}$ of 92.6\% and 95.4\%, respectively.  For {\it peopleother} (as shown in Fig. \ref{fig:PR_peopleother}), AP$^{0.75}$ is relatively low with a value of 59.2\%.  The top three effects on {\it peopleother} come from localization errors (19.6\%), class confusion within the {\it people} super-category (5.6\%) and false negatives (12.9\%).  The results stem from the definition rule for the {\it peopleother} class.  The definitions of the first five {\it people} classes (i.e. {\it pedestrian}, {\it cyclist}, {\it scooter}, {\it skater}, {\it sitter}) are very clear with distinctive features while {\it peopleother} is classified as a severely truncated or occluded class that cannot be recognized as any one of the five {\it people} classes.  Hence, the lack of explicitly defined features causes trouble in object localization and distinguishing the correct class.

\subsection{Behavioral Mapping Application}
\label{section:exp_mapping}

     \begin{figure} [tp]
        \vspace{-2.2 em}   
        \centering
        \begin{subfigure}[b]{0.4\textwidth}
            \centering
            \includegraphics[width=\textwidth]{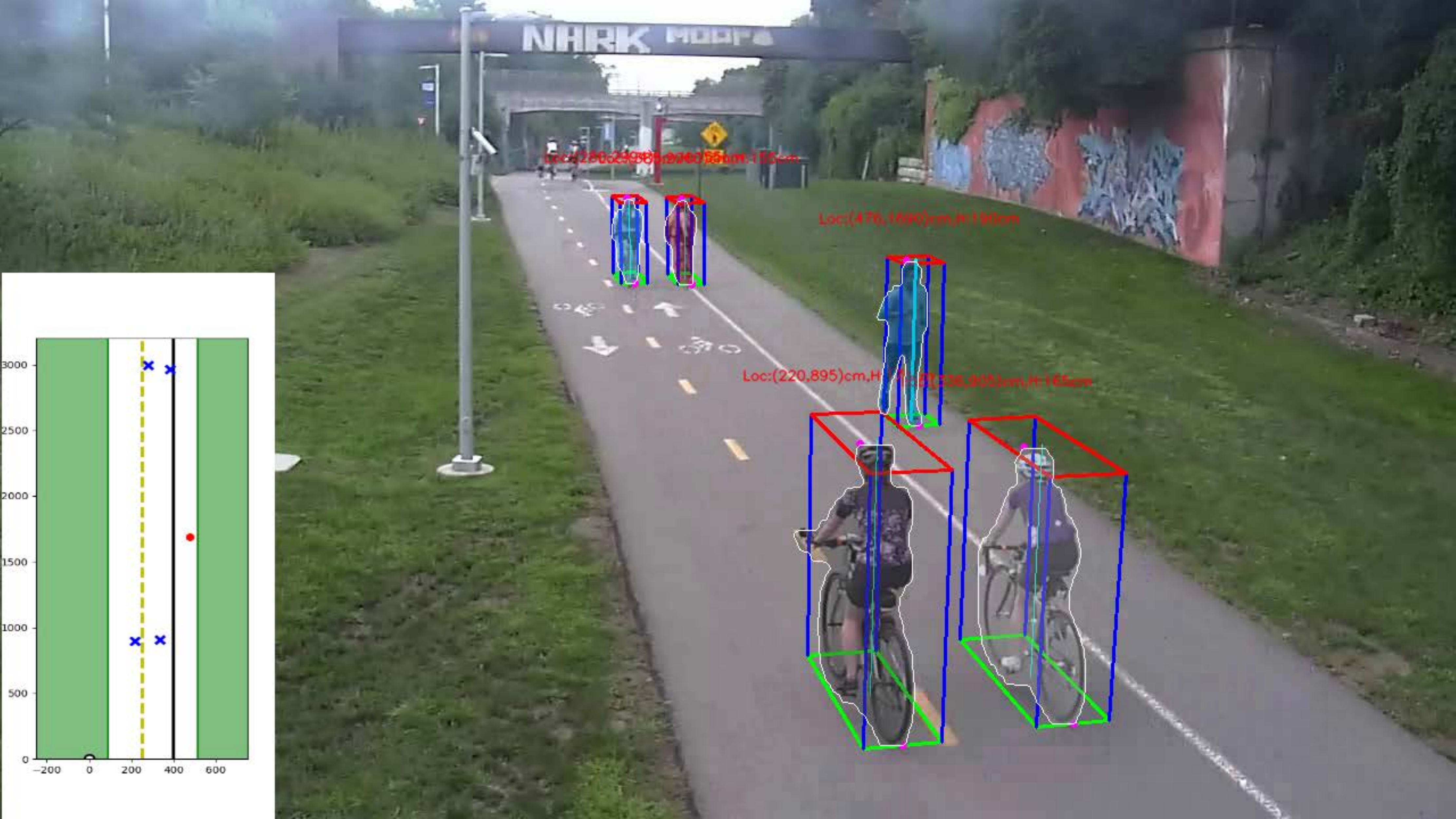}
        \end{subfigure}

        \caption{An example of user detection and mapping results on the Dequindre Cut. {\it pedestrian} is noted as {\color{red} $\bullet$}, {\it cyclist} is denoted as {\color{blue}x}, top and bottom pixels are denoted by {\color{magenta} $\boldsymbol{\cdot}$}.}
        \label{fig:dc_results}
        \vspace{-1.5em}   
    \end{figure}

  The locations of different users along the Dequindre Cut (a pedestrian path in the riverfront park) are projected to road maps (width=4.5m, length=32m) as shown in Fig. \ref{fig:dc_results}. The location and height of the 3D bounding boxes are estimated by using detected locations of object parts (e.g. feet, heads, bicycle tire) and \textit{a priori} defined horizontal sizes of pedestrians (w=50cm, l=60cm) and cyclists (w=50cm, l=160cm). It is found that 3D bboxes of moving {\it pedestrian} and {\it cyclist} can be estimated by using monocular cameras with some assumptions (e.g. flat ground and presumed geometry). 
 
     \begin{figure} [bp]
        \vspace{-1.8em}   
        \centering
        \begin{subfigure}[b]{0.23\textwidth}
            \centering
            \includegraphics[width=\textwidth]{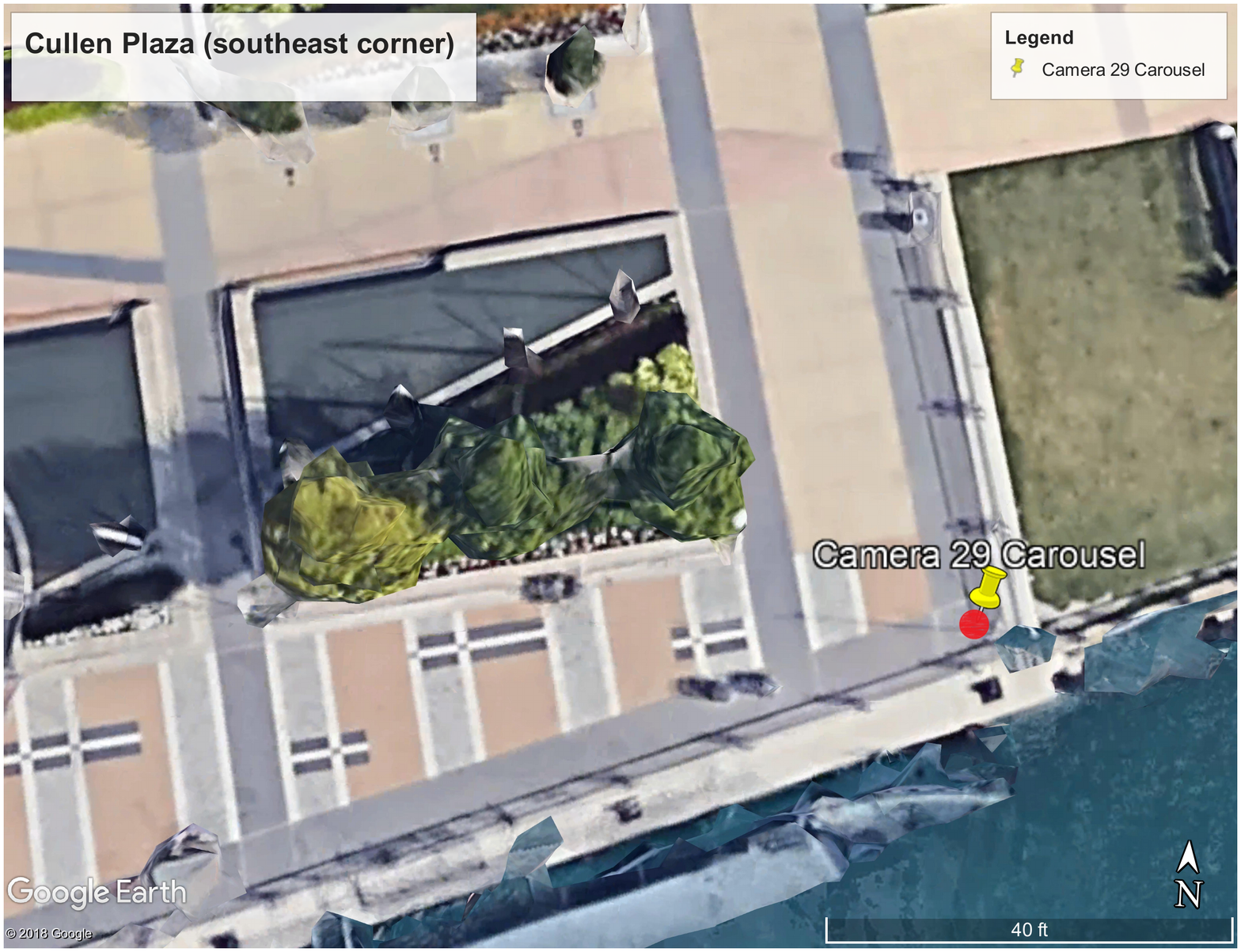}
            \vspace{-1.8em}   
            \caption[]%
            {{\footnotesize map with camera location}}    
            \label{fig:cam29_google}
        \end{subfigure}
        \,
        \begin{subfigure}[b]{0.23\textwidth}  
            \centering 
            \includegraphics[width=\textwidth]{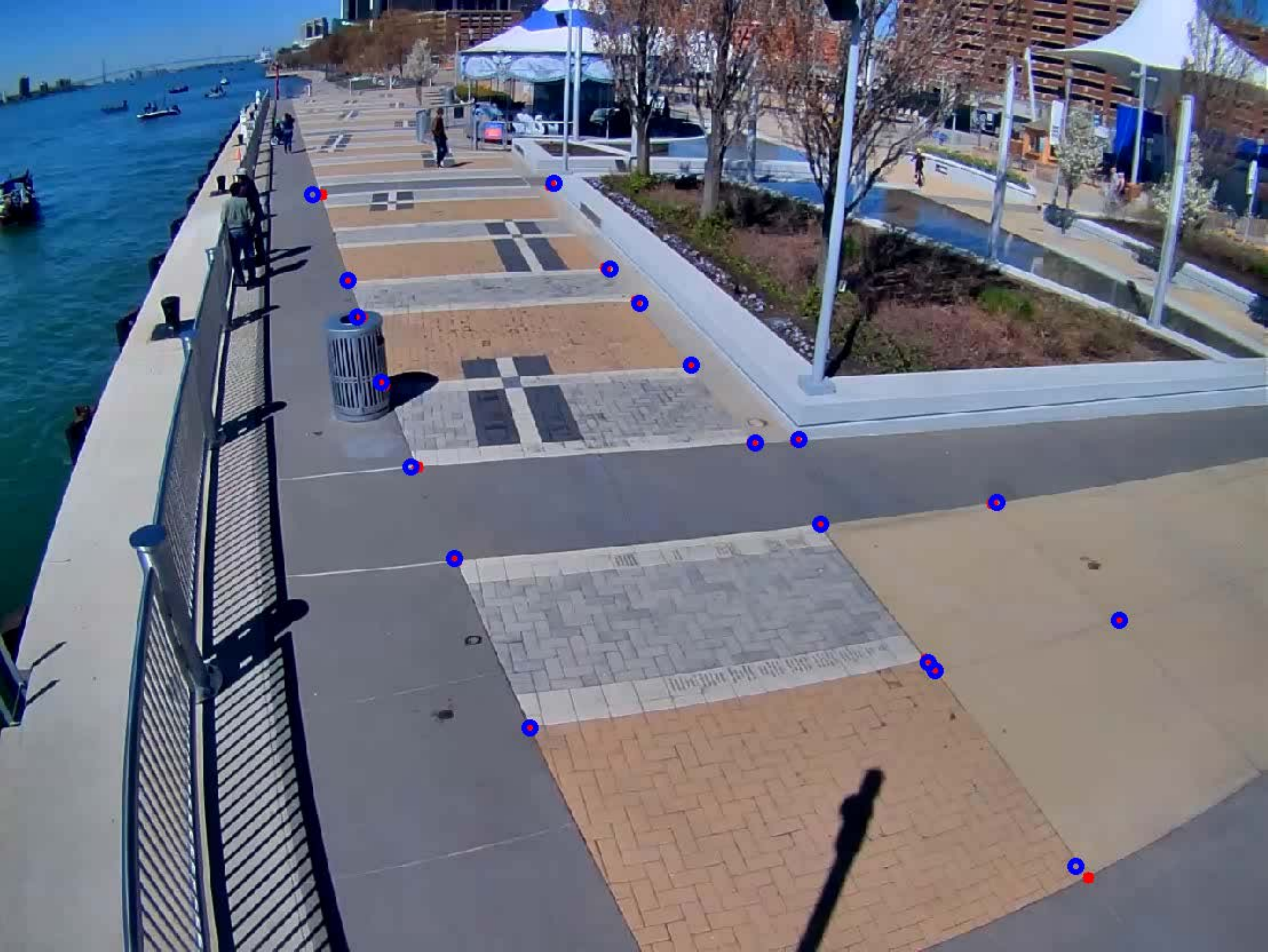}
            \vspace{-1.8em}   
            \caption[]%
            {{\footnotesize image with reference points}}    
            \label{fig:cam29_ref}
        \end{subfigure}

        \caption{Map of Cullen Plaza and camera calibration with reference points. {\color{red} $\bullet$}: selected pixels on image, {\color{blue} $\circ$}: projection of the world locations onto pixel coordinates using camera parameters.}
        \label{fig:cullen_results}
        \vspace{-2.2em}   
    \end{figure}
 
\begin{figure} [tp]
    \vspace{-2.2em}   
    \centering
    \begin{subfigure}[c]{0.23\textwidth}
    \centering
    \includegraphics[width=\textwidth]{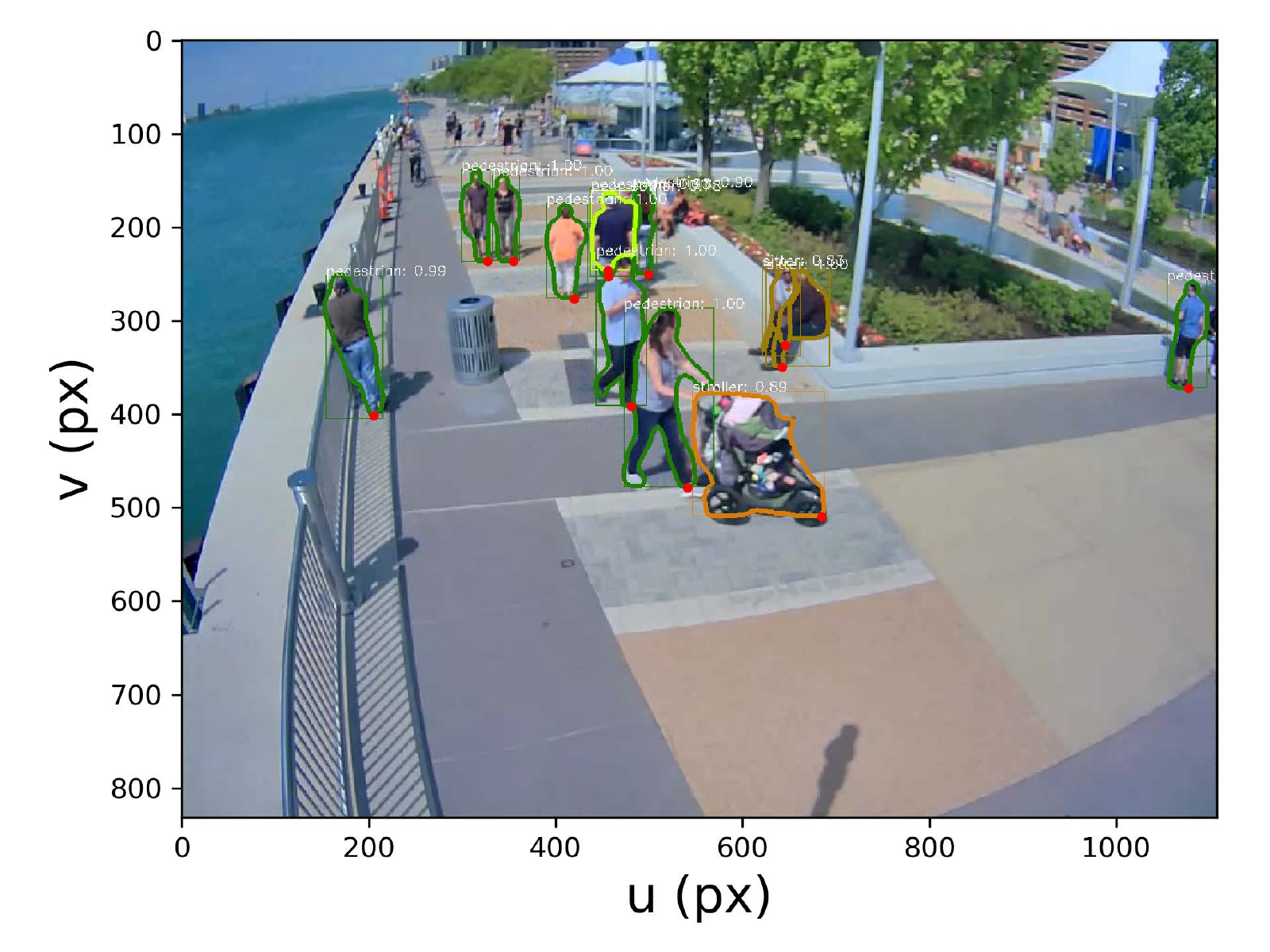}
    \vspace{-2.0em}   
    \caption[]%
        {{\footnotesize user detection}}    
    \label{fig:demo_cam29_detection}
    \end{subfigure}
    \,
    \begin{subfigure}[d]{0.23\textwidth}  
    \centering 
        \includegraphics[width=\textwidth]{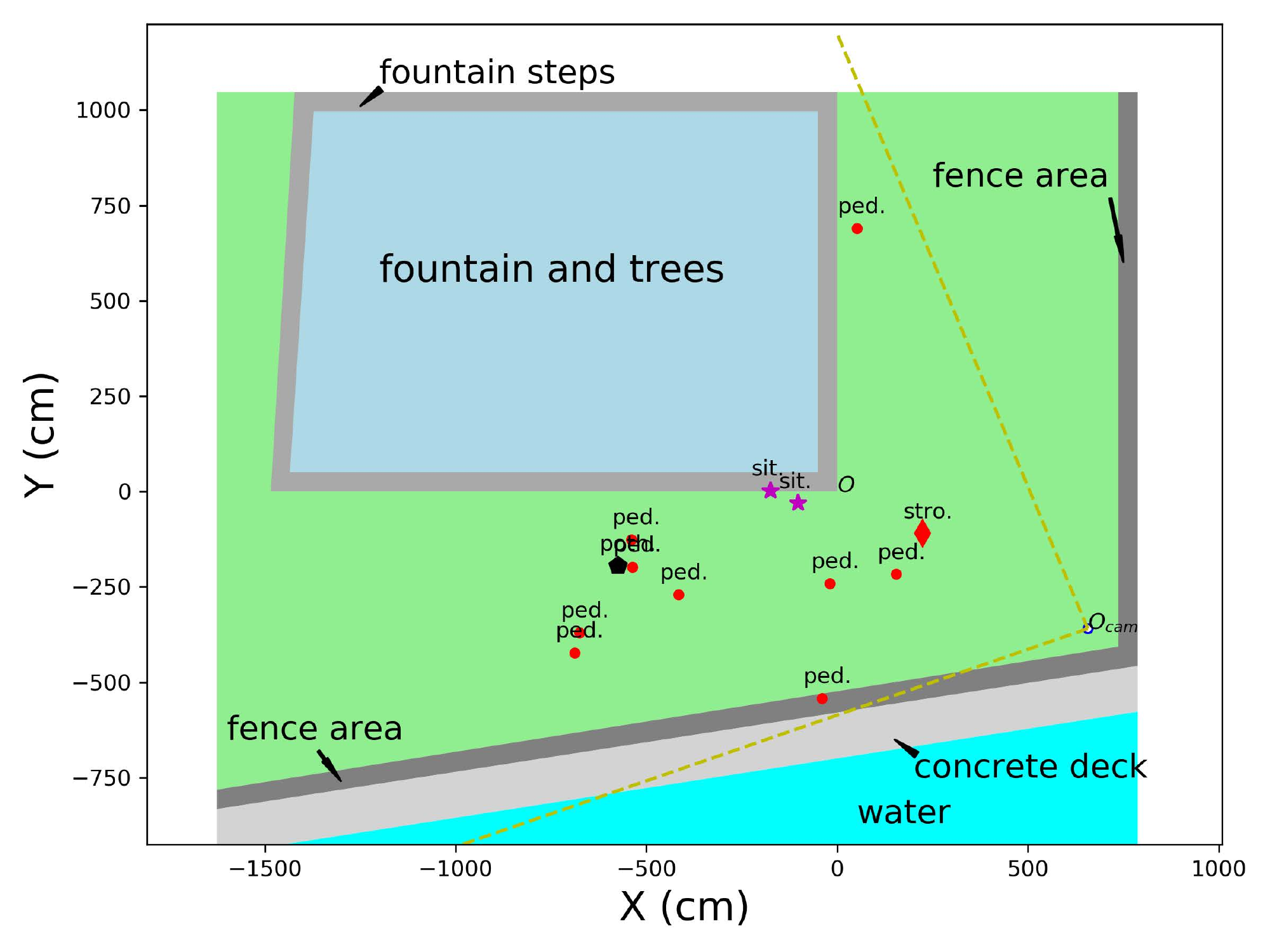}
    \vspace{-2.0em}   
    \caption[]%
        {{\footnotesize behavioral mapping}}    
    \label{fig:demo_cam29_mapping}
    \end{subfigure}
    \\
    
    \begin{subfigure}[c]{0.23\textwidth}
    \centering
        \includegraphics[width=\textwidth]{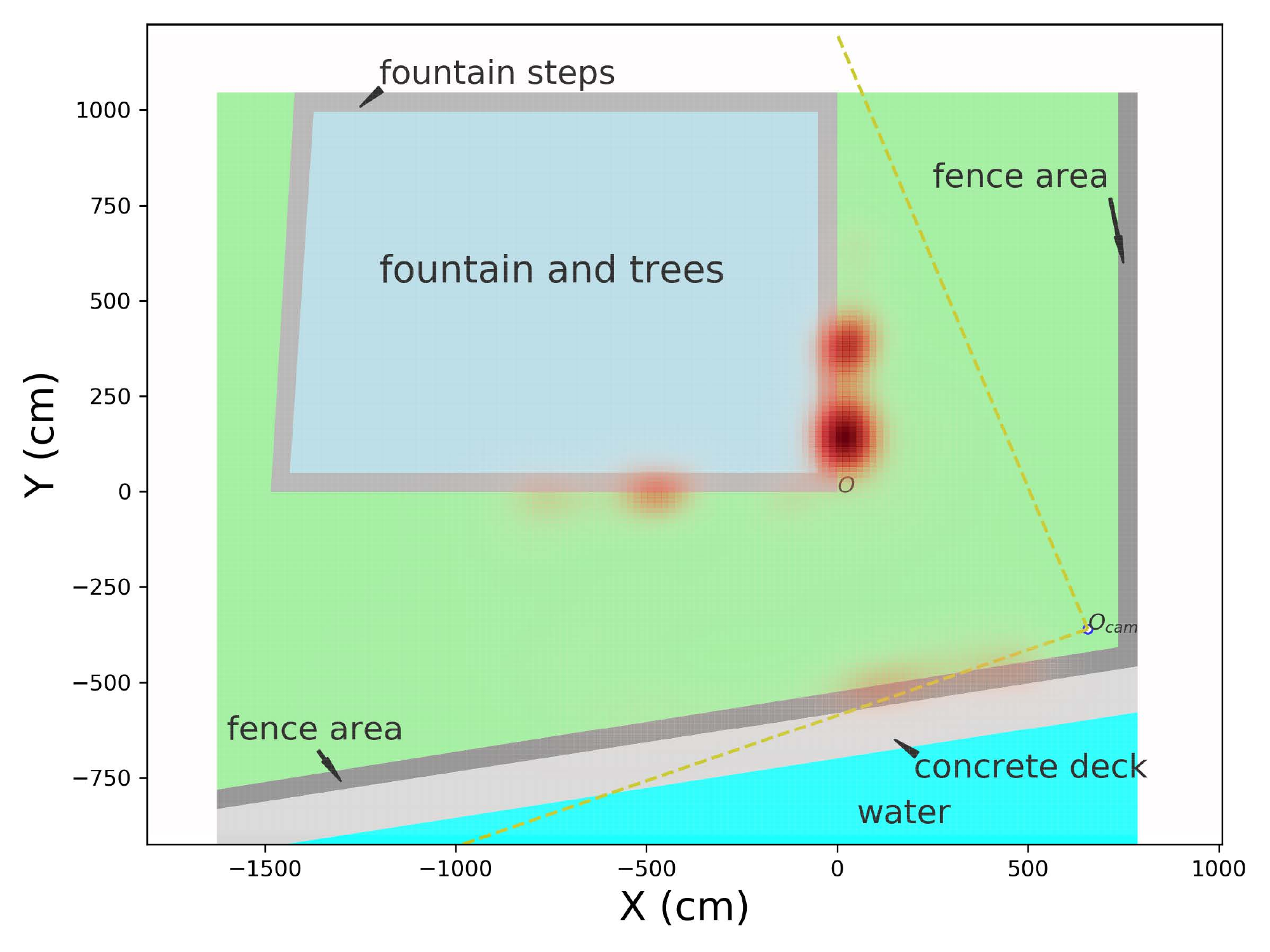}
    \vspace{-2.0em}   
    \caption[]%
        {{\footnotesize density map (day, 70.2k ct.)}}    
    \label{fig:kde_map_day}
    \end{subfigure}
    \,
    \begin{subfigure}[d]{0.23\textwidth}  
    \centering 
        \includegraphics[width=\textwidth]{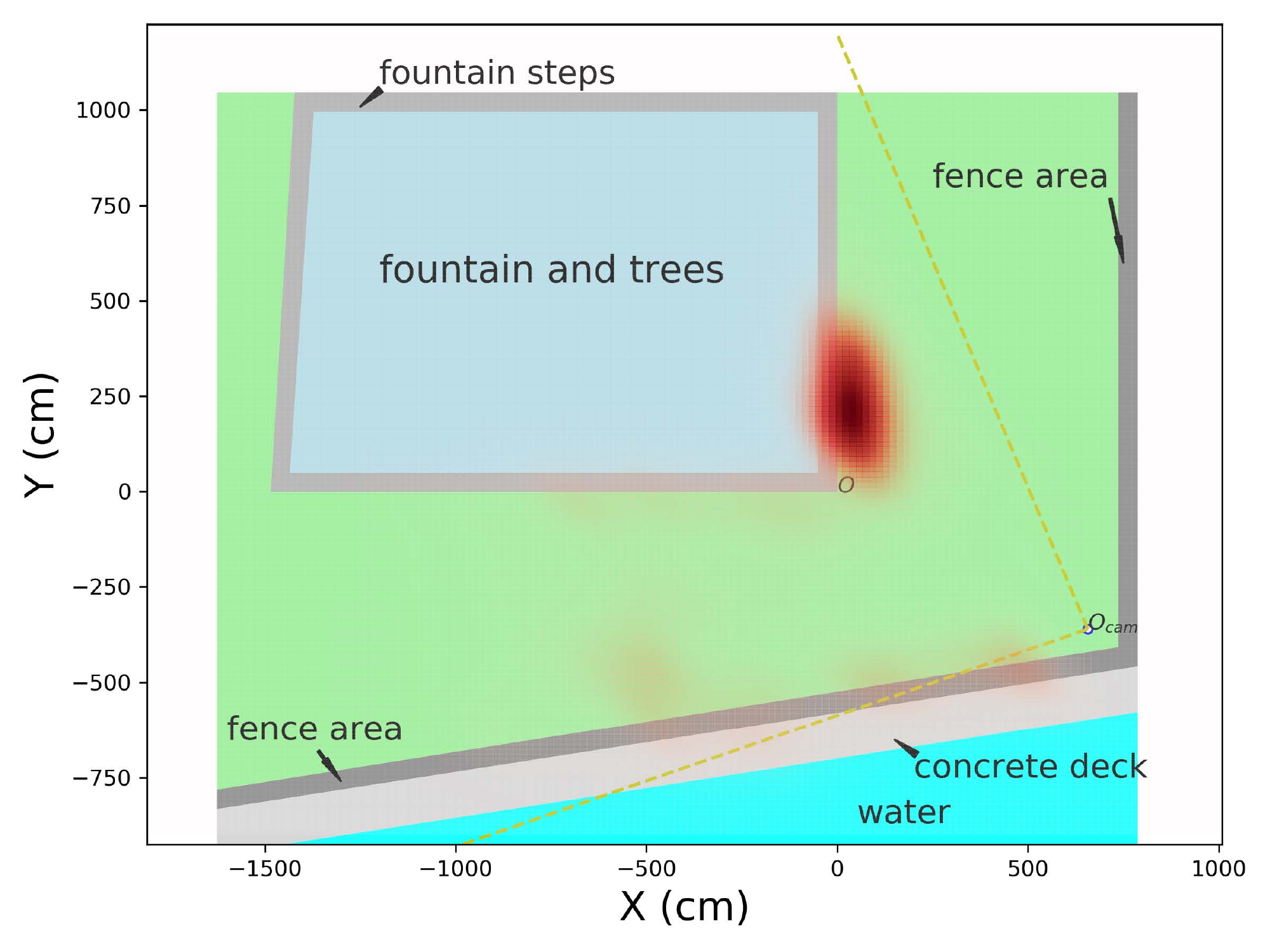}
    \vspace{-2.0em}   
    \caption[]%
        {{\footnotesize density map (week, 466.5k ct.)}}    
    \label{fig:kde_map_week}
    \end{subfigure}
    \\
    
    \caption{Examples of (a) user detection, (b) behavioral mapping results,  (c) daily density map, and (d) weekly density map on the southeast corner of Cullen Plaza.}
    \label{fig:plaza_results}
    \vspace{-1.8em}   
\end{figure}

A camera at Cullen Plaza (Fig. \ref{fig:cam29_google}) is calibrated by using a checkerboard and 19 reference points on the ground (shown in Fig. \ref{fig:cam29_ref}). The error of the mapping is obtained by computing the difference between field measurements of the reference points and the projection of the corresponding image pixels. The averaged error for a static scene is 7.68cm, which is accurate enough for urban design studies in a large POS. User localization (Fig. \ref{fig:demo_cam29_detection}) and behavioral mappings (Fig. \ref{fig:demo_cam29_mapping}) at any specific moment can be obtained in 0.16 s/frame. However, a statistical study over a period (e.g. density map) can provide more insights to understand the utilization of POS.

Long-term monitoring of users ({\it people} detections are filtered) is achieved by generating density maps using accumulated detection data over space and time. The density maps over one day (e.g. 2019-06-25, 9:00am to 17:00, in Fig. \ref{fig:kde_map_day}) and one week (e.g. from 2019-06-24 to 2019-06-30, 9:00 to 17:00, in Fig. \ref{fig:kde_map_week}) are generated by using kernel density estimation \cite{silverman2018density} on the detection results (frames are sampled at 1 fps). It is found that users tend to stay closer to the edges (e.g. fountain steps, poles, fence area, statues, etc.) in a POS. For example, {\it sitter} tend to sit on fountain steps and {\it pedestrian} tend to stay near fence poles. The scenario is consistent with the ``edge effect'' that has been observed by architects and urban planners in the past \cite{gehl2013cities}.

\section{Conclusion} \label{section: conclusion}
In this paper, the OPOS dataset is presented along with a baseline detection model of Mask R-CNN. The custom dataset is specifically designed for user detection in POS using surveillance cameras. A benchmark study of user sensing and behavioral mapping at the Detroit Riverfront Conservancy (DRFC) is demonstrated, providing general guidelines in building a CV-based framework to measure usage of POS for many urban design applications. The detection results show that the baseline detector has a bbox mAP of 68.4\% (67.0\% in segmentation) for overall objects, and a bbox mAP of 70.9\% (70.5\% in segmentation) for the {\it people} super-category. The bbox AP$^{0.75}$ for the most two common people classes ({\it pedestrian} and {\it cyclist}) that appear at the Detroit Riverfront are 92.6\% and 95.4\%, respectively. The averaged error for the behavioral mapping task is 7.68 cm which is satisfactory for urban studies in large public spaces. To spur future research on CV-based measurement of POS usage, evaluation metrics and an error diagnosis method is also provided to analyze the detection models.  In the end, a case study of the proposed method is performed to measure the utilization of a popular plaza at the Detroit Riverfront in a week.   In the future, the study would serve as a stepping stone to other challenging tasks (e.g. user counting, tracking, re-id tasks) that are associated with urban planning studies.

\section*{Acknowledgement}
The support from the National Science Foundation (NSF) under grant \#1831347 is gratefully acknowledged.
\newpage

{\small
\bibliographystyle{ieee}
\bibliography{WACV2020_ps}
}

\end{document}